\documentclass[a4paper]{article}

\usepackage{amssymb,amsmath,amsfonts,graphicx}
\usepackage[english]{babel}
\usepackage{theorem}

\def\R{{\mathbb{R}}}

\def\Div{\textup{div\,}}
\newtheorem{theoreme}{Th\'eor\`eme}[section]
\newtheorem{lemme}[theoreme]{Lemma}

\newtheorem{definition}{Definition}
\newtheorem{proposition}[theoreme]{Proposition}

\title{Noisy image decomposition: a new structure, texture and noise model based on local adaptivity}
\author{J\'er\^ome Gilles}

\date{}

\begin{document}
\maketitle

\begin{center}
Centre d'Expertise Parisien \\ 16 bis, avenue Prieur de la C\^ote d'Or 94114 ARCUEIL Cedex FRANCE\\
jerome.gilles@etca.fr\\ The original publication is available at www.springerlink.com.
\end{center}

\begin{abstract}
These last few years, image decomposition algorithms have been proposed to split an image into two parts: the structures and the textures. These algorithms are not adapted to the case of noisy images because the textures are corrupted by noise. In this paper, we propose a new model which decomposes an image into three parts (structures, textures and noise) based on a local regularization scheme. We compare our results with the recent work of Aujol and Chambolle. We finish by giving another model which combines the advantages of the two previous ones.\\


\end{abstract}

\section{Introduction}
In 2001, in \cite{meyer}, Y.Meyer investigated the model of image restoration proposed by Rudin-Osher-Fatemi \cite{rof}:
\begin{equation} \label{eqn:rof}
F^{ROF}(u)=J(u)+(2\lambda)^{-1}\|f-u\|_{L^2}^2
\end{equation}
where $f$ is the measured image which implies an estimation: 
\begin{equation}
\hat{u}=\arg\inf_{u\in BV}F^{ROF}(u),
\end{equation}
of the restored image. The main hypothesis is that $u$ belongs to the space $BV$ (the space of bounded variation functions). The quantity $J(u)$ is a semi-norm on $BV$. It is also knowing as the total variation of the function $u$ and can be expressed by
\begin{equation}
J(u)=\int |\nabla u|.
\end{equation}
If we write $v=f-u$ and take the point of view of image decomposition (\emph{i.e} $f$ is composed of structures ($u$) and textures ($v$)), we can rewrite equation (\ref{eqn:rof}) as
\begin{equation}
F^{ROF}(u)=J(u)+(2\lambda)^{-1}\|v\|_{L^2}^2.
\end{equation}

Y.Meyer proved that this model rejects the oscillatory component of $f$ which is considered to be the textured component. Then, he proposed a new model where the $L^2$ norm of $v$ is replaced by a norm in a space $G$ close to the dual space of $BV$ and which contains oscillatory functions.
The new model introduced by Y.Meyer is then given by
\begin{equation}
F^{YM}(u,v)=J(u)+(2\lambda)^{-1}\|v\|_G,
\end{equation}
where the $G-$norm is defined by:
\begin{definition}
For $v=\partial_1 g_1+\partial_2 g_2$ where $g_1\in L^{\infty}(\mathbb{R}^2),g_2\in L^{\infty}(\mathbb{R}^2)$,
\begin{equation}\label{eq:normeg}
\|v\|_G=\inf_{g}\left\|\left(\left|g_1\right|^2+\left|g_2\right|^2\right)^{\frac{1}{2}}\right\|_{L^{\infty}}.
\end{equation}
\end{definition}

Due to its non-linearity, the $G-$norm is difficult to compute numerically (see \cite{meyer}).
Then two approaches can be found in the litterature to run Meyer's algorithm.\\
The first one is the algorithm proposed by L.Vese and S.Osher in \cite{vese1}. The authors use the following approximation:
\begin{equation}\label{eqn:vesehyp}
\forall v \in L^{\infty} \quad \|v\|_{L^{\infty}}=\lim_{p\rightarrow +\infty}\|v\|_{L^p},
\end{equation}
and the property that any function $v$ in $G$ can be written as $v=\Div (g)$ where $g=(g_1,g_2)\in (L^{\infty}\times L^{\infty})$. Then they give a new formulation of the problem:
\begin{equation}\label{eqn:vese}
F^{VO}(u,g)=J(u)+ (2\lambda)^{-1} \|f- (u+\Div \; g)\|_{L^2}^2+ \mu \left\|\sqrt{g_1^2+g_2^2}\right\|_{L^p}.
\end{equation}
The authors seek for
\begin{equation}
(\hat{u},\hat{g_1},\hat{g_2})=\arg\inf_{(u,g_1,g_2)\in (BV\times L ^{\infty}\times L^{\infty}}F^{VO}(u,g_1,g_2))
\end{equation}

The authors deduce the related partial differential equation system and its discrete formulation (see \cite{vese1} for details). The drawback of their approach is the numerical stability of the algorithm.\\

\begin{figure}[t!]
\begin{center}
\begin{tabular}{ccc}
\includegraphics[width=0.3\textwidth]{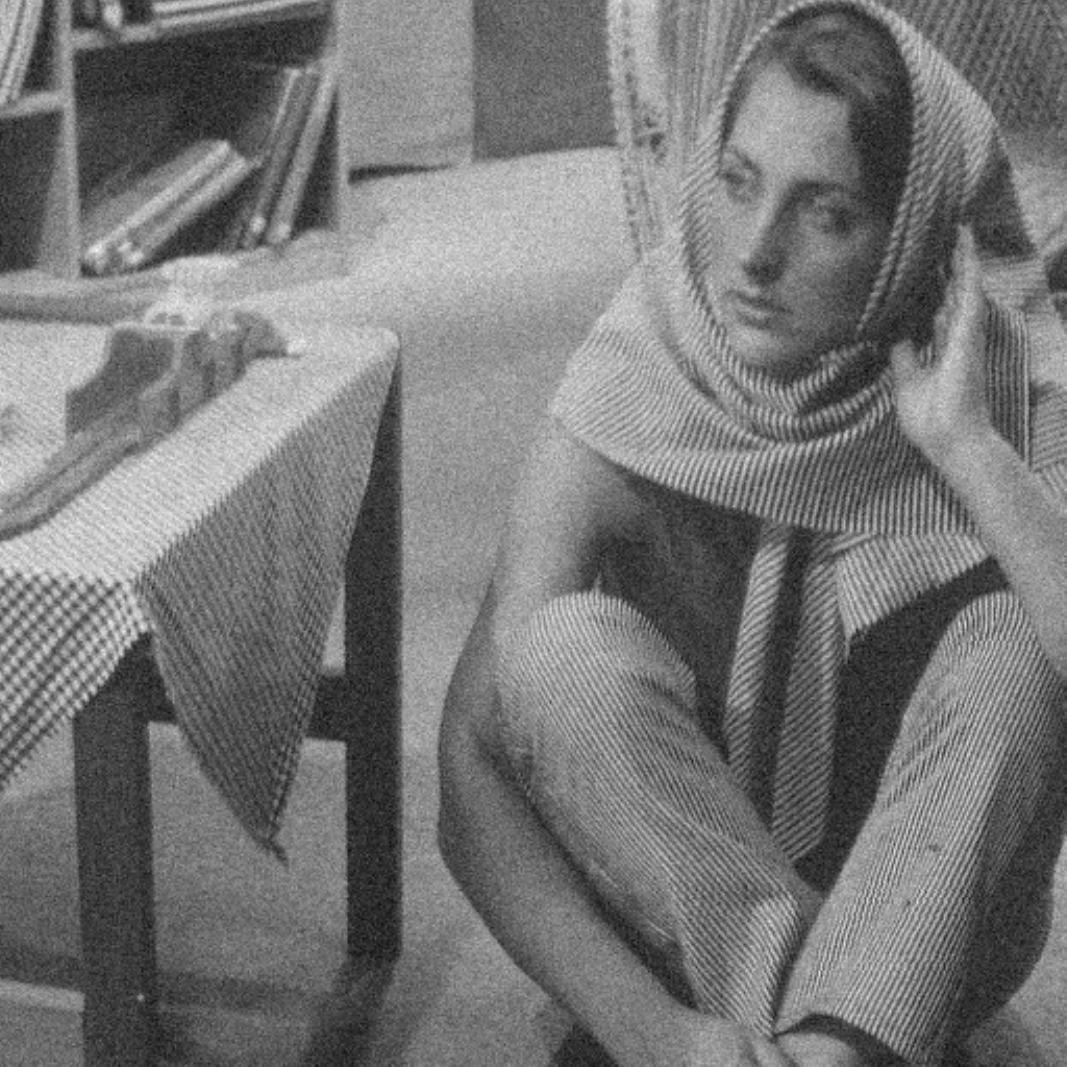} &
\includegraphics[width=0.3\textwidth]{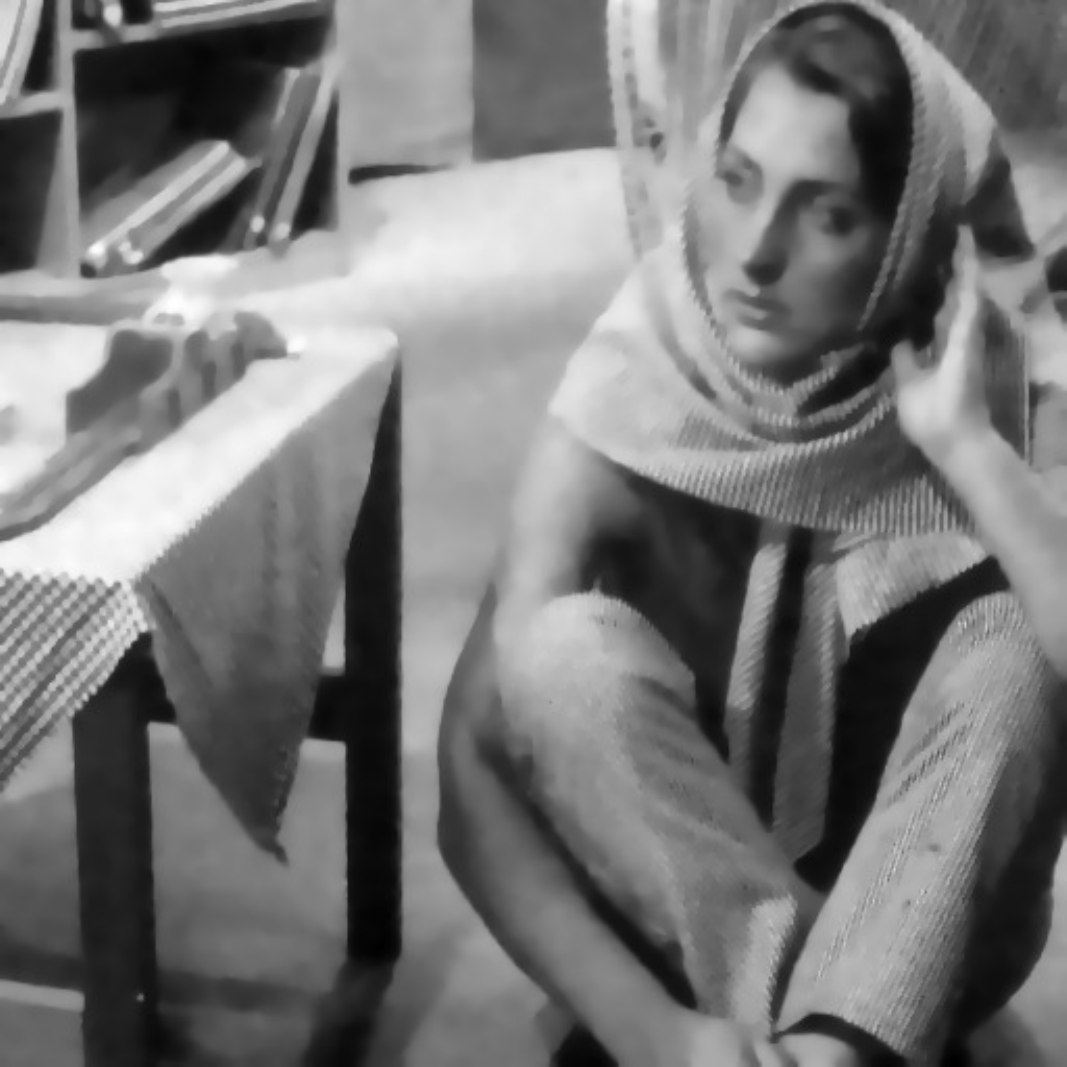} &
\includegraphics[width=0.3\textwidth]{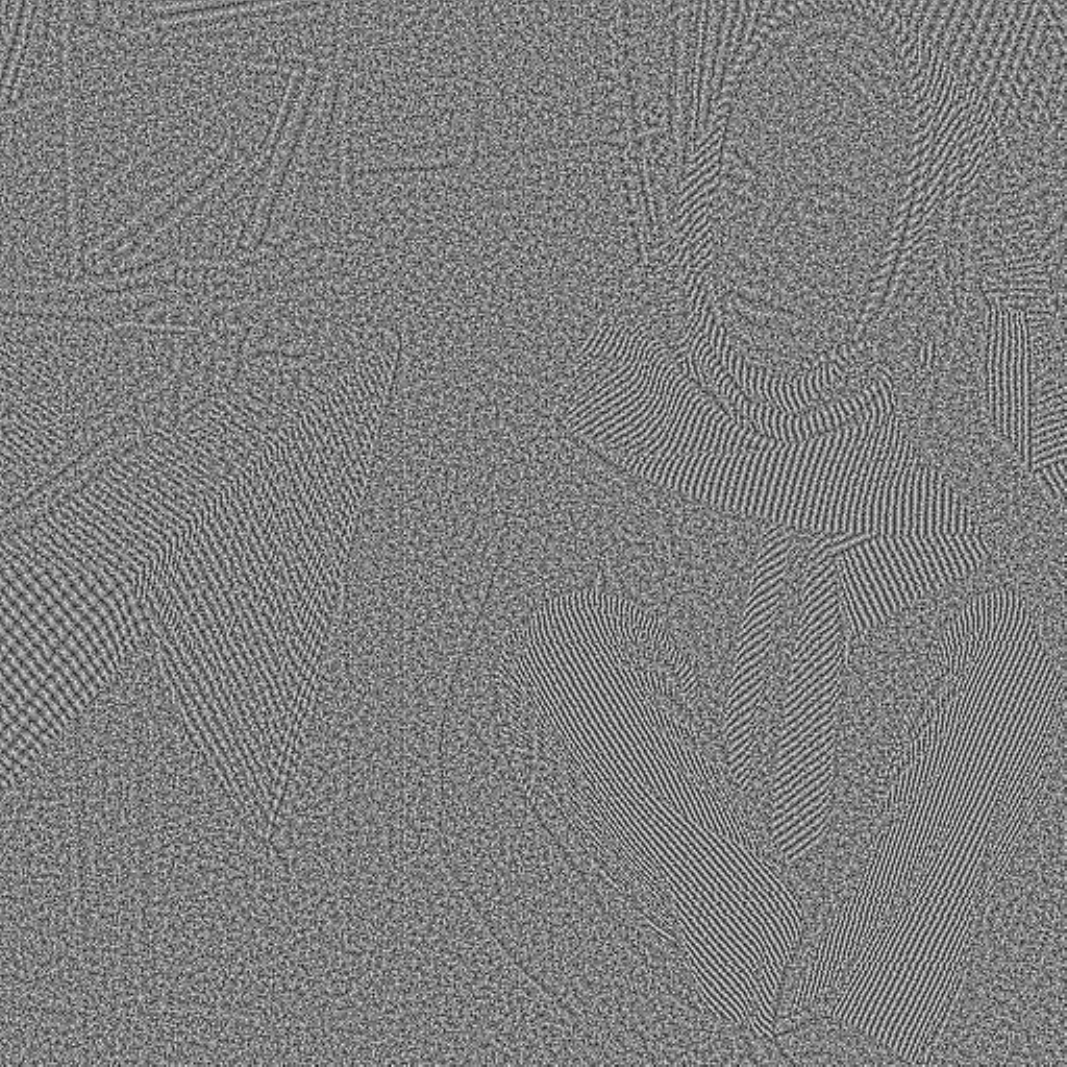} 
\end{tabular}
\caption{Two part decomposition of noisy image, from left to right: noisy image, object part, texture part corrupted by noise.}
\label{fig:vnoise}
\end{center}
\end{figure}

The second approach is proposed by Aujol et al. in \cite{aujol,aujolphd}. Their model is given by
\begin{equation}\label{eq:aujoluv}
F^{AU}(u,v)=J(u)+ J^*\left(\frac{v}{\mu}\right)+(2\lambda)^{-1}\|f-u-v\|_{L^2}^2,
\end{equation}
\begin{equation}
(\hat{u},\hat{v})=\arg\inf_{(u,v)\in (BV\times G}F^{AU}(u,v))
\end{equation}
where
\begin{equation}
G_{\mu}=\{v\in G/\|v\|_G\leqslant \mu\},
\end{equation}
and $J^*$ is the adjoint of $J$; therefore it is the indicator function defined by
\begin{equation}\label{eq:indicator}
J^*\left(\frac{v}{\mu}\right)=
\begin{cases}
&0 \quad \quad \text{if} \; v\in G_{\mu},\\
&+\infty \; \;\; \text{if} \; v\in G \backslash G_{\mu}.
\end{cases}
\end{equation}
The authors prove that the solution of minimizing (\ref{eq:aujoluv}) can be found by an iterative algorithm based on non-linear projectors $P_{G_{\mu}}$ proposed by A.Chambolle (see \cite{chambolle} for details and for a convergence theorem).
This algorithm gives good results but is of limited interest in the case of noisy images. Indeed, the noise can be viewed as a very highly oscillatory function (this means that noise is in $G_{\mu}$). Therefore the algorithm incorporates the noise in the texture component. Then the textures are corrupted by noise (see fig.\ref{fig:vnoise} for an example).\\

In this paper, we propose to extend the two components model to a three components model, $f=u+v+w$, which discriminates between structures ($u$), textures ($v$) and noise ($w$). This work was initiated in \cite{jegilles}.

An outline of the paper is given now. In section \ref{sec:uvwlocal}, we describe our new algorithm based on a local regularization principle \cite{gilboa,aujol2}. We give some numerical aspects of the algorithm and show some results. In section \ref{sec:aujol} we recall the recent model proposed by Aujol and Chambolle \cite{aujol-chambolle} which also decomposes an image into three parts. A comparison between our approach and their model shows the advantages and disadvantages of each algorithm. In section \ref{sec:adaptwav}, we propose another algorithm which incorporates the advantages of each the previous ones. We conclude and give some perspectives of this work in section \ref{sec:conclusion}.

\section{Three part image decomposition: a local adaptative algorithm}
\label{sec:uvwlocal}
In this section, we propose a new model to decompose an image into three parts: structures ($u$), textures ($v$) and noise ($w$). As in the $u+v$ model, we consider that structures and textures are modelized by functions in $BV$ and $G$ spaces respectively. We also consider a zero mean gaussian noise added to the image. Let us view noise as a specific very oscillating function. In virtue to the work in \cite{meyer}, where it is shown that the more a function is oscillatory, the smaller its $G-$norm is, we propose to modelize $w$ as a function in $G$ and consider that its $G-$norm is much smaller than the norm of the textures ($\|v\|_G  \gg \|w\|_G$). These assumptions are equivalent to choose
\begin{equation}
v\in G_{\mu_1}\text{ , }w\in G_{\mu_2} \qquad \text{where} \quad \mu_1 \gg \mu_2
\end{equation}

In order to increase the performance, we propose to add a local adaptability behaviour to the algorithm following an idea proposed by G.Gilboa et al. in \cite{gilboa}. The authors investigate the ROF model given by equation (\ref{eqn:rof}) and propose a modified version which is able to preserve textures in the denoising process. To do this, they don't choose $\lambda$ as a constant on the whole image but as a function $\lambda(f)(x,y)$ which represents local properties of the image. In a cartoon-type region, the algorithm enhances the denoising process by increasing the value of $\lambda$; in a texture-type region the algorithm decreases $\lambda$ to attenuate the regularization in order to preserve the details of textures. So $\lambda(f)(x,y)$ can be viewed as a smoothed partition between textured and untextured regions.\\
Then, in order to decompose an image into three parts, we propose to use the following functionnal:

\begin{equation}
F_{\lambda ,\mu_1 ,\mu_2}^{JG}(u,v,w)= J(u)+J^*\left(\frac{v}{\mu_1}\right)+J^*\left(\frac{w}{\mu_2}\right)+(2\lambda)^{-1}\|f-u-\nu_1 v-\nu_2 w\|_{L^2}^2,
\end{equation}

where the functions $\nu_i$ represent the smoothed partition of textured and untextured regions (and play the role of $\lambda$ in Gilboa's paper). The $\nu_i$ functions must have the following behaviour:\\
\begin{itemize}
\item if we are in a textured region, we want to favour $v$ instead of $w$. This is equivalent to $\nu_1$ close to $1$ and $\nu_2$ close to $0$,
\item if we are in an untextured region, we want to favour $w$ instead of $v$. This is equivalent to $\nu_1$ closed to $0$ and $\nu_2$ close to $1$.\\
\end{itemize}
We see that $\nu_1$ and $\nu_2$ are complementary so it is natural to choose $\nu_2=1-\nu_1:\mathbb{R}^2\rightarrow ]0;1[$. The choice of $\nu_1$ and $\nu_2$ is discussed after the following proposition which characterizes the minimizers of $F_{\lambda ,\mu_1 ,\mu_2}^{JG}(u,v,w)$.

\begin{proposition}\label{prop:uvw}
Let $u\in BV$, $v\in G_{\mu_1}$, $w\in G_{\mu_2}$ be the structures, textures and noise parts respectively and $f$ the original noisy image. Let the functions $(\nu_1(f)(.,.),\nu_2(f)(.,.))$ be defined on $\mathbb{R}^2\rightarrow ]0;1[$, and assume that these functions could be considered as locally constant compared to the variation of $v$ and $w$ (see the proof for details). Then a minimizer defined by
\begin{equation}\label{equ:uvw2}
(\hat{u},\hat{v},\hat{w})=\underset{(u,v,w)\in BV\times G_{\mu_1} \times G_{\mu_2}}{\arg}\min F_{\lambda ,\mu_1 ,\mu_2}^{JG}(u,v,w),
\end{equation}
is given by
\begin{align}
\hat{u}&=f-\nu_1\hat{v}-\nu_2\hat{w}-P_{G_{\lambda}}(f-\nu_1\hat{v}-\nu_2\hat{w}), \\
\hat{v}&=P_{G_{\mu_1}}\left(\frac{f-\hat{u}-\nu_2\hat{w}}{\nu_1}\right), \\
\hat{w}&=P_{G_{\mu_2}}\left(\frac{f-\hat{u}-\nu_1\hat{v}}{\nu_2}\right),
\end{align}
where $P_{G_{\mu}}$ is the Chambolle's non-linear projectors (see \cite{chambolle}).\\
\end{proposition}

The proof of this proposition is given in appendix \ref{ann:a}. Notice that we have still no results on the uniqness of the minimizer.\\

In pratice, we propose the following iterative numerical scheme:

\begin{enumerate}
\item initialization: $u_0=v_0=w_0=0$,
\item compute $\nu_1$ and $\nu_2=1-\nu_1$ from $f$ (see below for the choice of $\nu_i$),
\item compute $w_{n+1}=P_{G_{\mu_2}}\left(\frac{f-u_n-\nu_1v_n}{\nu_2+\kappa}\right)$, ($\kappa$ is a small value in order to prevent the division by zero),
\item compute $v_{n+1}=P_{G_{\mu_1}}\left(\frac{f-u_n-\nu_2w_{n+1}}{\nu_1+\kappa}\right)$,
\item compute $u_{n+1}=f-\nu_1v_{n+1}-\nu_2w_{n+1}-P_{G_{\lambda}}(f-\nu_1v_{n+1}-\nu_2w_{n+1})$,
\item if $\max\{|u_{n+1}-u_n|,|v_{n+1}-v_n|,|w_{n+1}-w_n|\}\leqslant\epsilon$ or if we did $N_{step}$ \\ iterations then stop the algorithm, else jump to step 3.\\
\end{enumerate}

Now, let's deal with the choice of the $\nu_i$ functions. In \cite{gilboa}, the authors choose to compute a local variance on the texture + noise part of the image obtained by the ROF model ($f-u$). In this paper, we use the same strategy but on the $v$ component obtained by the two part decomposition algorithm. This choice is implied by the fact that the additive gaussian noise can be considered orthogonal to the textures. As a consequence, the variance of a textured region is larger than the variance of an untextured region.\\
So, in practice, we first compute the two part decomposition of the image $f$. On the $v$ component, $\forall (i,j)$, we compute the local variance on a small window (odd size $L$) centered on $(i,j)$. Then $\nu_1(i,j)$ can be expressed by
\begin{align}
\nu_1(i,j)&=\frac{1}{L^2}\sum_{(p,q)\in Z_{ij}}v^2(p,q)-\frac{1}{L^4}\left(\sum_{(p,q)\in Z_{ij}} v(p,q) \right)^2, \\
\nu_2(i,j)&=1-\nu_1(i,j),
\end{align}
where $Z_{ij}=[i-(L-1)/2;i+(L-1)/2]\times[j-(L-1)/2;j+(L-1)/2]$.

\begin{figure}[t!]
\begin{center}
\includegraphics[width=0.3\textwidth]{barb512noise} 
\includegraphics[width=0.3\textwidth]{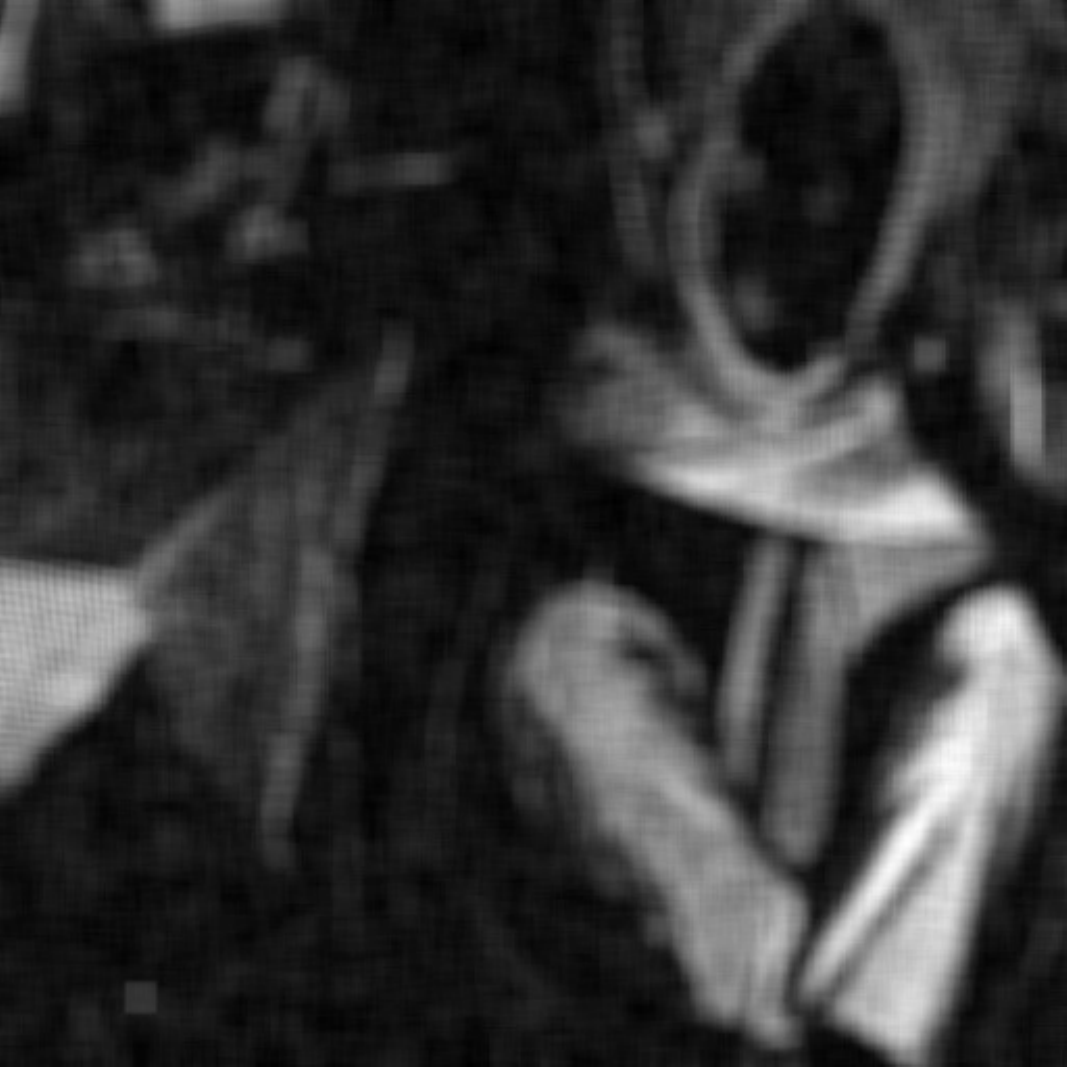}
\caption{On left we have the noisy image (Barbara$+$gaussian noise, $\sigma=20$); on right, texture partition $\nu_1$.}
\label{fig:nu}
\end{center}
\end{figure}

At least, we normalize it in order to get the values in $]0;1[$. Figure \ref{fig:nu} shows an example on a noisy version of Barbara ($\sigma=20$). We see that this image, which represents $\nu_1$, has larger values in the textured regions. Denote that in \cite{gilboa2}, the same authors propose to estimate the optimal parameter based on a SNR criteria. At this time, we did not incorporate this method in our algorithm.\\

\begin{figure}[t!]
\centering
\begin{tabular}{ccc}
\includegraphics[width=0.3\textwidth]{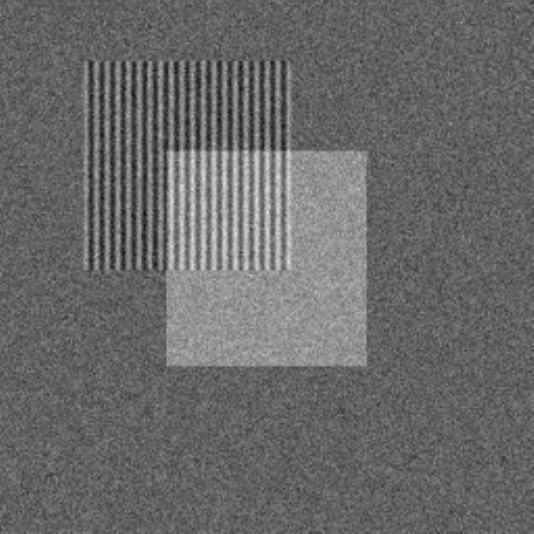} &  
\includegraphics[width=0.3\textwidth]{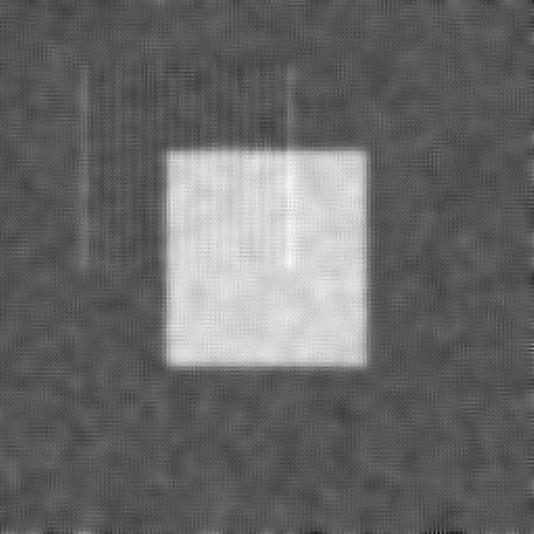} \\  
Noisy original image & Structures \\

\includegraphics[width=0.3\textwidth]{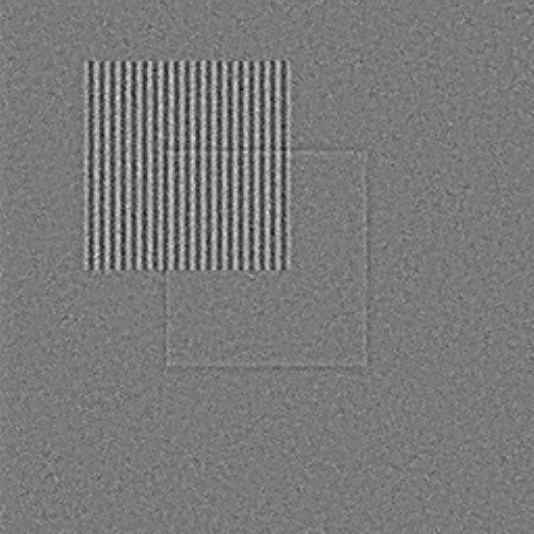} &
\includegraphics[width=0.3\textwidth]{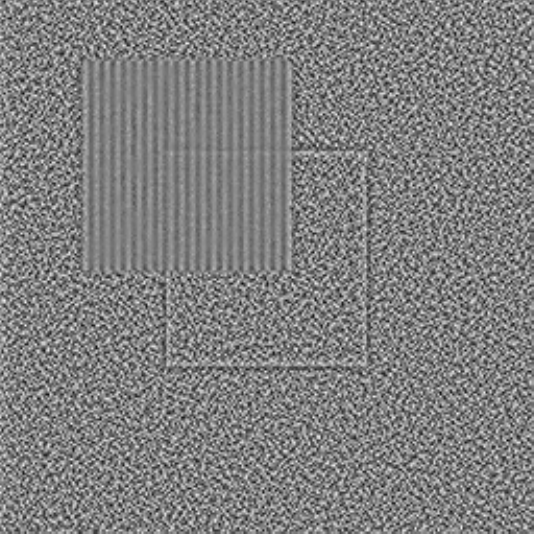} \\ 
Textures & Noise
\end{tabular}
\caption{Results given by $F_{\lambda, \mu_1 ,\mu_2}^{JG}$ applied on the synthetic image.}
\label{fig:resultscar}
\end{figure}

\begin{figure}[t!]
\hspace{-7mm}
\begin{tabular}{ccc}
\includegraphics[width=0.49\textwidth]{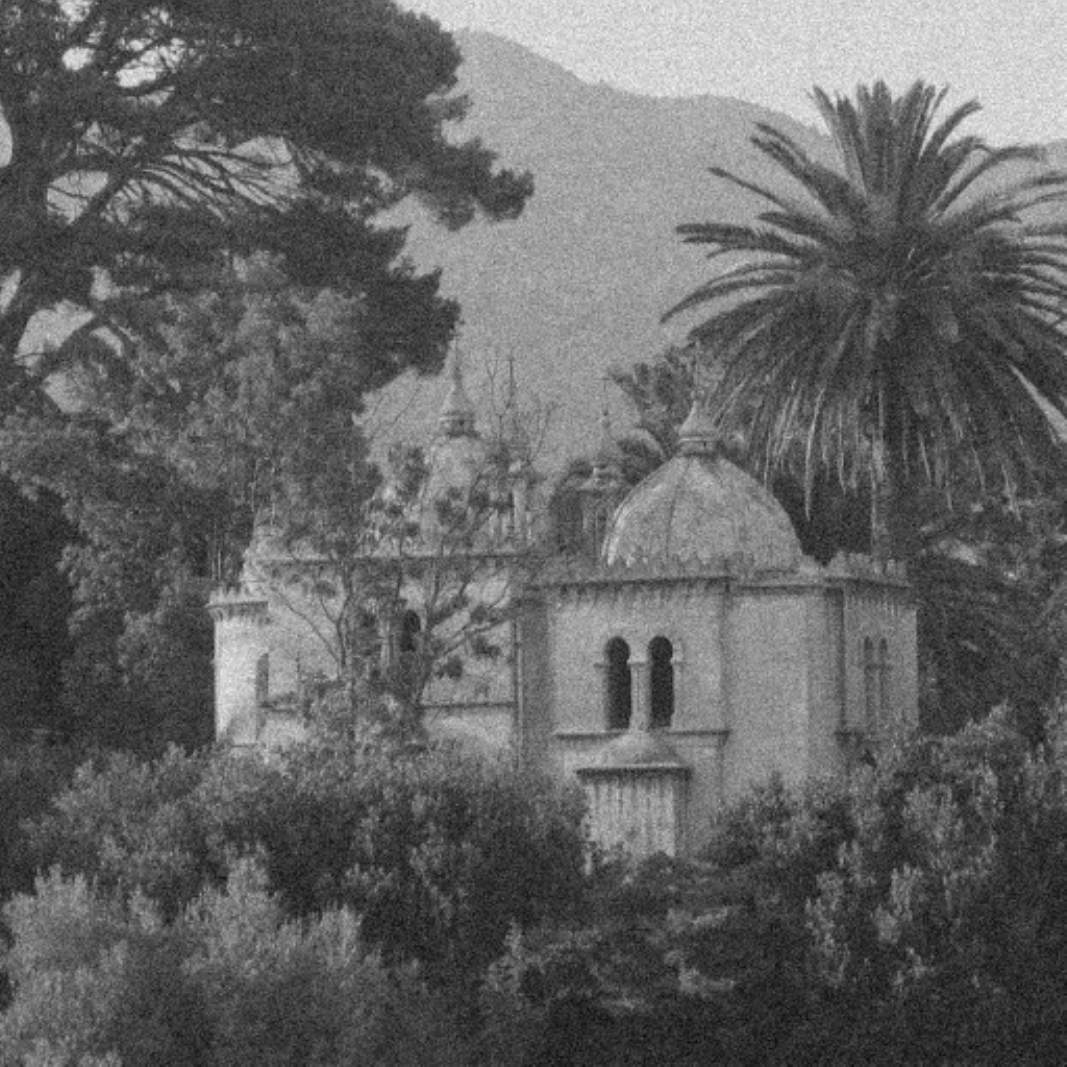} &  
\includegraphics[width=0.49\textwidth]{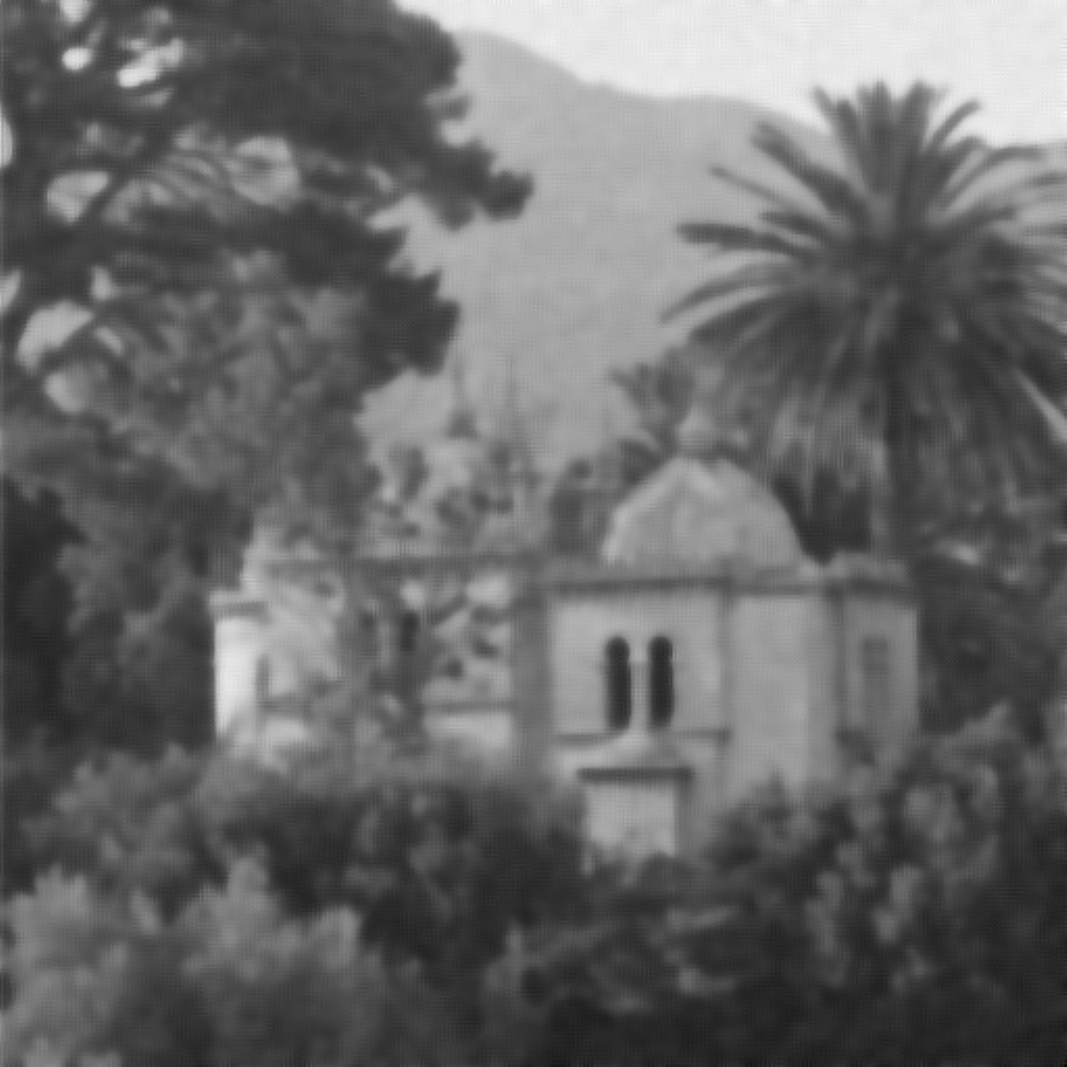} \\  
Noisy original image & Structures \\
\includegraphics[width=0.49\textwidth]{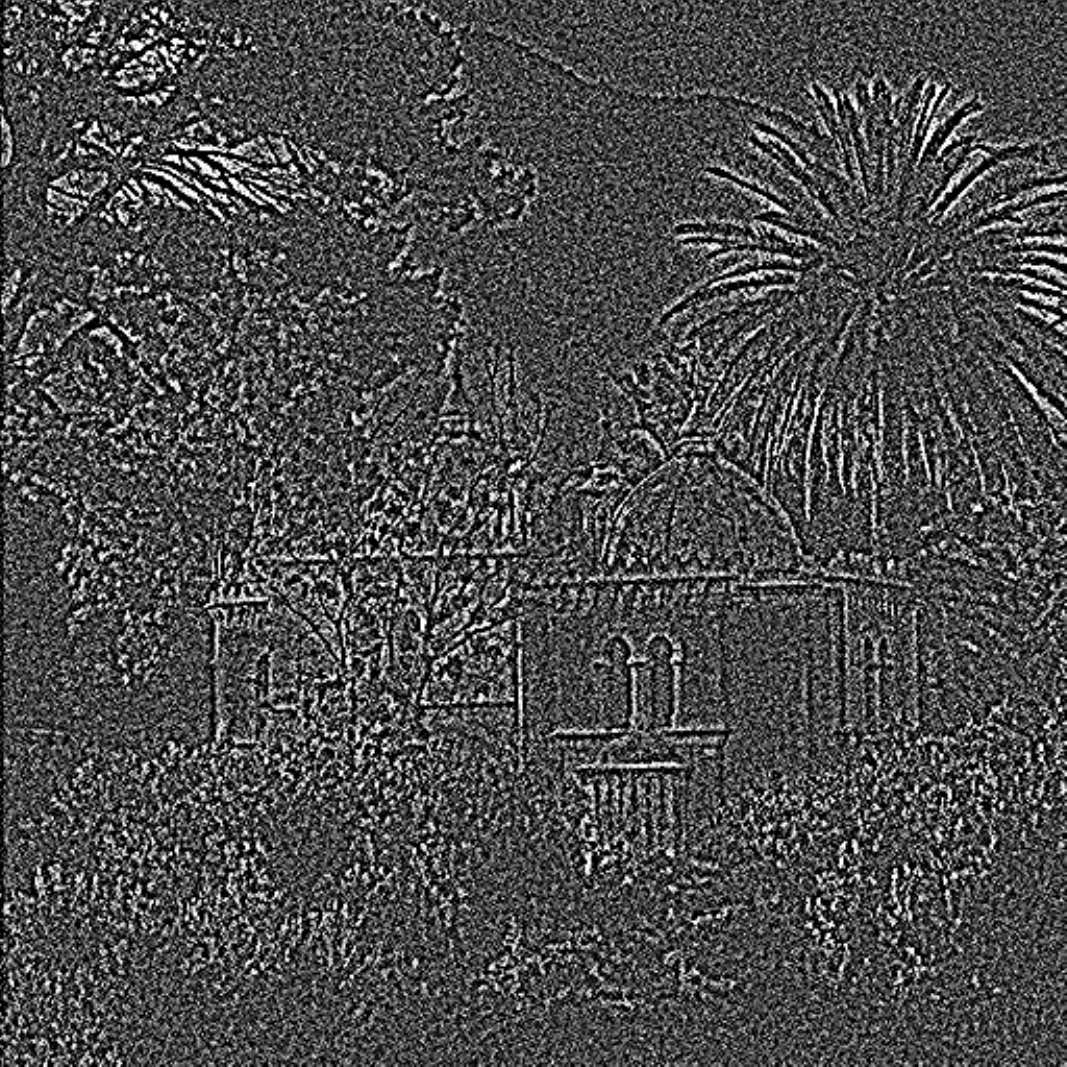} &
\includegraphics[width=0.49\textwidth]{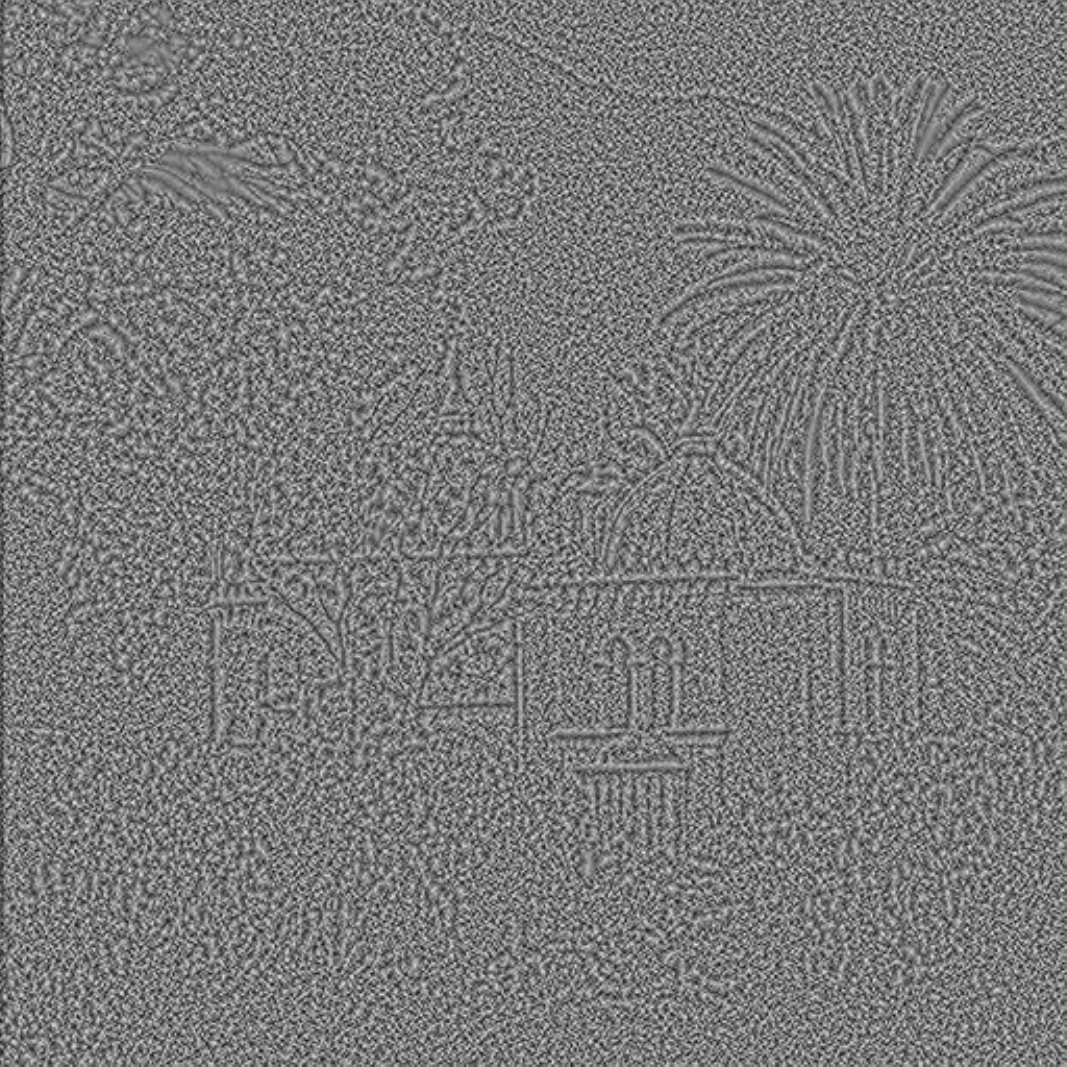} \\ 
Textures & Noise
\end{tabular}
\caption{Results given by $F_{\lambda, \mu_1 ,\mu_2}^{JG}$ applied on the real outdoor image.}
\label{fig:resultsbat}
\end{figure}

We experimented our algorithm on three different images: on a synthetic one (show on the top-left of figure \ref{fig:resultscar}), on Barbara (show on the left of figure \ref{fig:nu}) and on a real outdoor image (show on the top-left of figure \ref{fig:resultsbat}). The parameters are set to
\begin{itemize}
\item Synthetic: $\lambda=50$, $\mu_1=1000$, $\mu_2=10$, the size of the window for computing $\nu$ is 7 pixels,
\item Barbara: $\lambda=10$, $\mu_1=1000$, $\mu_2=1$, the size of the window for computing $\nu$ is 15 pixels,
\item Outdoor: $\lambda=50$, $\mu_1=1000$, $\mu_2=10$, the size of the window for computing $\nu$ is 7 pixels.\\
\end{itemize}

\begin{figure}[t!]
\hspace{-7mm}
\begin{tabular}{ccc}
\includegraphics[width=0.33\textwidth]{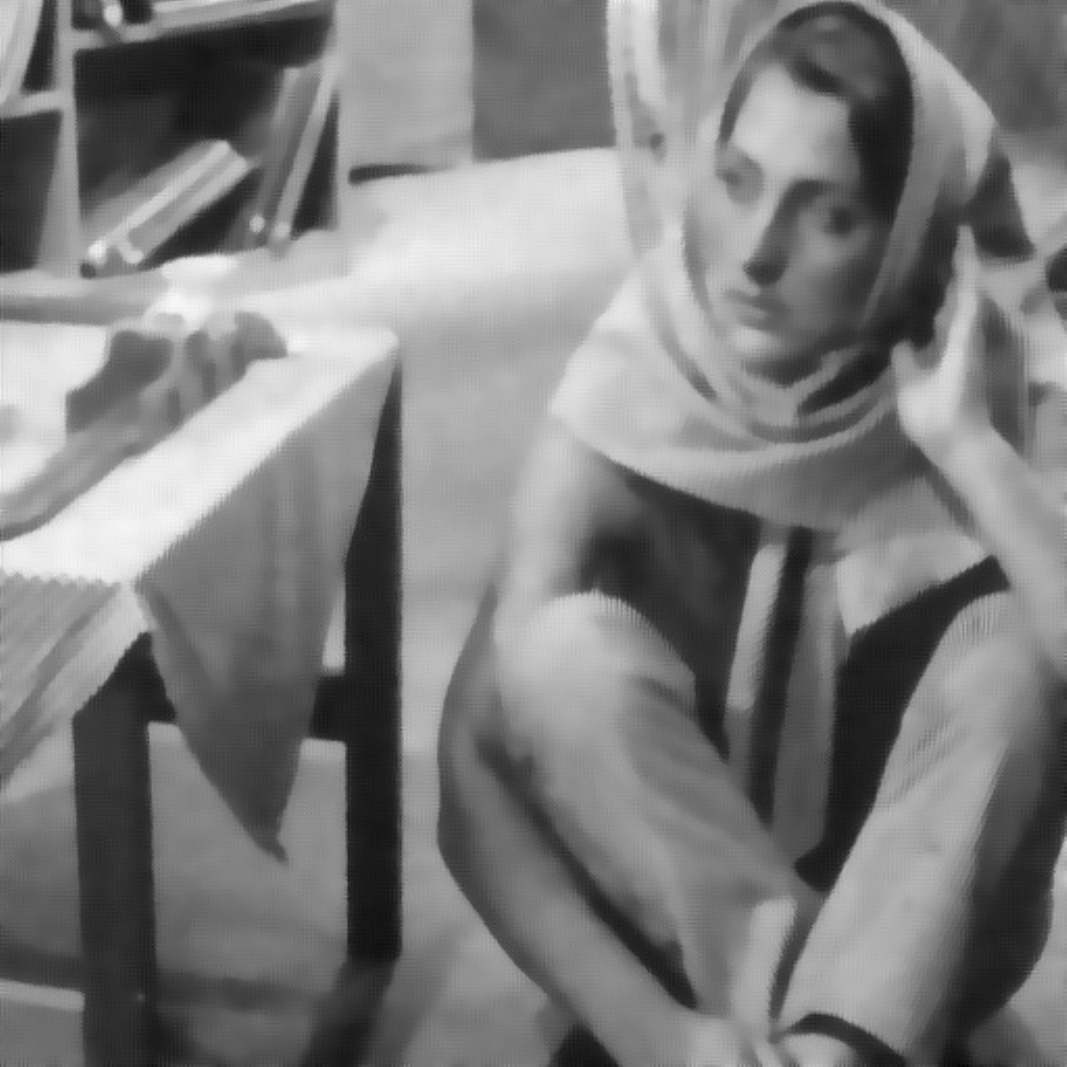} &  
\includegraphics[width=0.33\textwidth]{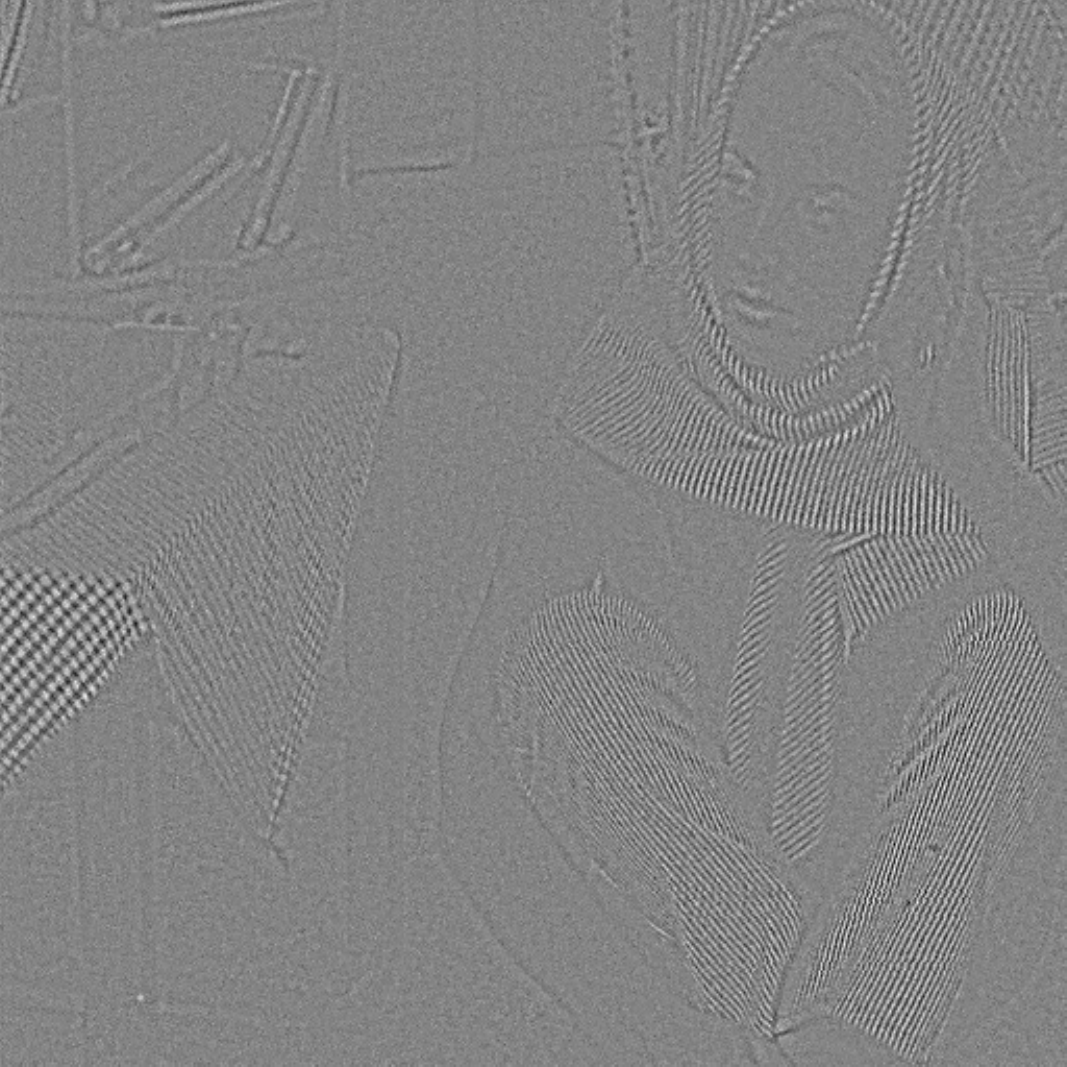} &
\includegraphics[width=0.33\textwidth]{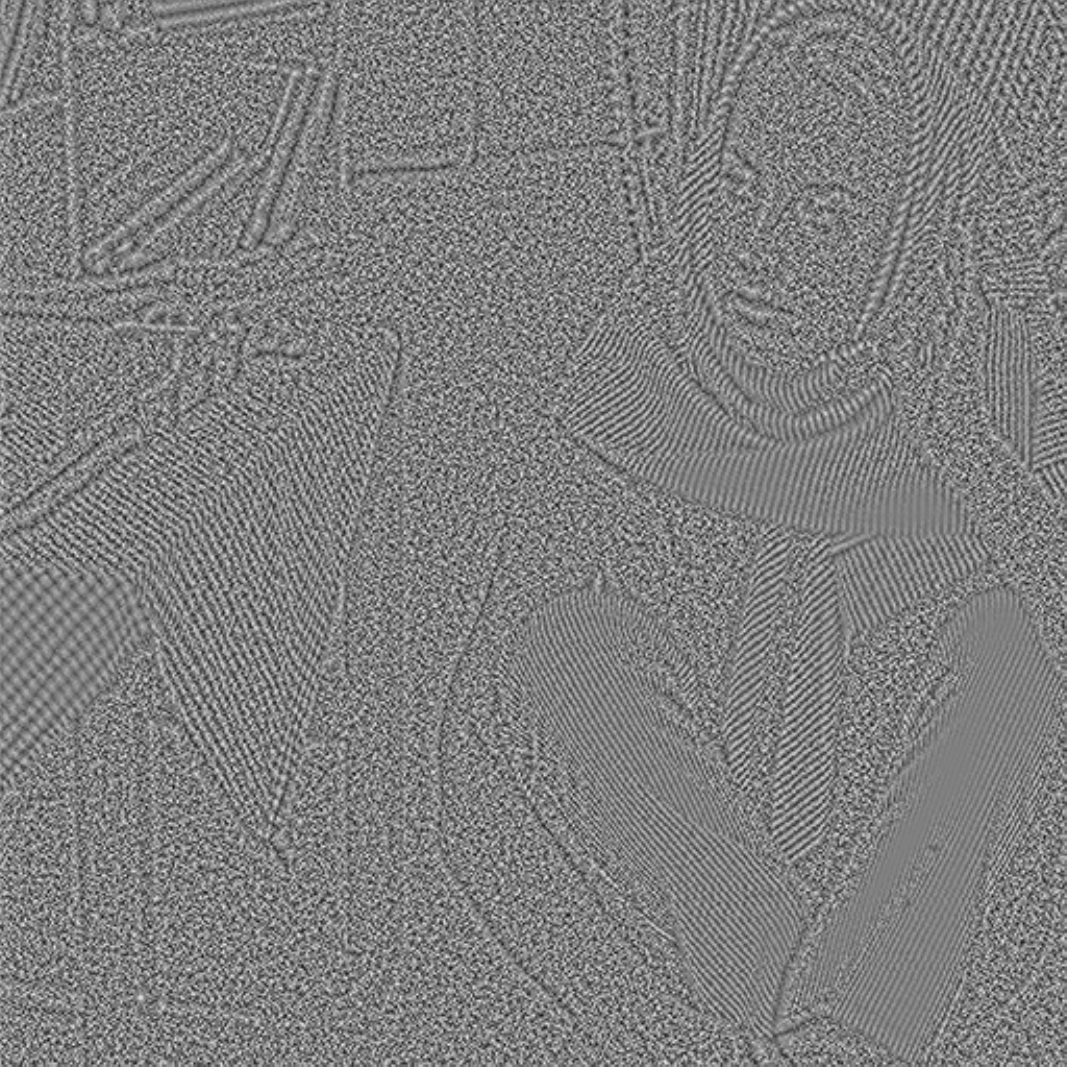} \\ 
Structures & Textures & Noise \\

\includegraphics[width=0.33\textwidth]{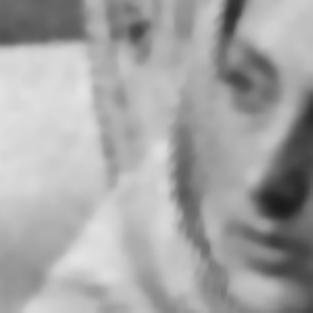} &  \includegraphics[width=0.33\textwidth]{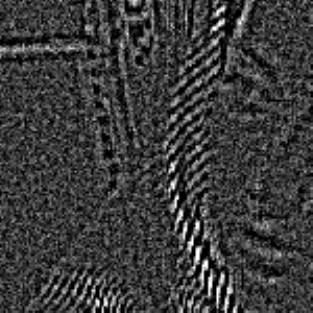} &
\includegraphics[width=0.33\textwidth]{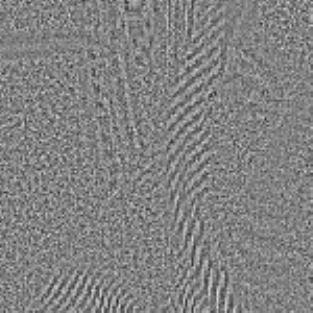} \\ 
Structures zoom & Textures zoom & Noise zoom

\end{tabular}
\caption{Results given by $F_{\lambda, \mu_1 ,\mu_2}^{JG}$ applied on Barbara. First row gives the whole images and second row zooms on a section of the images.}
\label{fig:resultsbarb}
\end{figure}

The results are given respectively in figures \ref{fig:resultscar}, \ref{fig:resultsbarb} and \ref{fig:resultsbat}. We can see in the whole cases, that the proposed algorithm successfully separates the noise from textures. In the zoomed images of figure \ref{fig:resultsbarb}, we can see that some residual noise stayed in the textured part and some residual texture is extracted in the noise part. This is inevitable since the $\nu_i$ functions made a tradeoff between extracting the information in the texture part or in the noise part.

\section{Aujol-Chambolle's algorithm}\label{sec:aujol}
In \cite{aujol-chambolle}, the authors propose another model to decompose an image into three parts. Their assumption is to consider noise as a distribution modelized by the Besov space $\Dot{B}^{\infty}_{-1,\infty}$, and the following functional is used:
\begin{equation}\label{eqn:aujoluvw}
F_{\lambda,\mu,\delta}^{AC}(u,v,w)= J(u)+J^*\left(\frac{v}{\mu}\right)+B^*\left(\frac{w}{\delta}\right)+(2\lambda)^{-1}\|f-u-v-w\|_{L^2}^2,
\end{equation}
where $u\in BV$,$v\in G_{\mu}$, $w\in E_{\delta}$ are defined by (we use the standard setting $s=-1,p=q=+\infty$ for the Besov space $B_{s,q}^p$):
\begin{equation}
E_{\delta} =\left\{w\in \Dot{B}^{\infty}_{-1,\infty} / \|w\|_{\Dot{B}^{\infty}_{-1,\infty}}\leqslant \delta\right\}.
\end{equation}

The term $B^*(.)$ is the indicator function on $E_{\delta}$ defined in the same way as equation (\ref{eq:indicator}). Then the minimizer is given by (see \cite{aujol-chambolle}):
\begin{align}
&\hat{u}=f-\hat{v}-\hat{w}-P_{G_{\lambda}}(f-\hat{v}-\hat{w}),\\
&\hat{v}=P_{G_{\mu}}(f-\hat{u}-\hat{w}),\\
&\hat{w}=P_{E_{\delta}}(f-\hat{u}-\hat{v})=f-\hat{u}-\hat{v}-WST(f-\hat{u}-\hat{v},2\delta),
\end{align}
where $WST(f-\hat{u}-\hat{v},2\delta)$ is the Wavelet Soft Thresholding operator applied on $f-\hat{u}-\hat{v}$ with a threshold set to $2\delta$ ($\delta=\eta\sigma\sqrt{\log N}$ and $N$ is the number of pixels, $\eta$ is a tuning parameter, see \cite{donohoseuil,chambolle2}). A numerical iterative algorithm can be deduced from these formula (see \cite{aujol-chambolle}).\\

\begin{figure}[t!]
\centering
\begin{tabular}{ccc}
\includegraphics[width=0.3\textwidth]{carre_texture_noisy} &  
\includegraphics[width=0.3\textwidth]{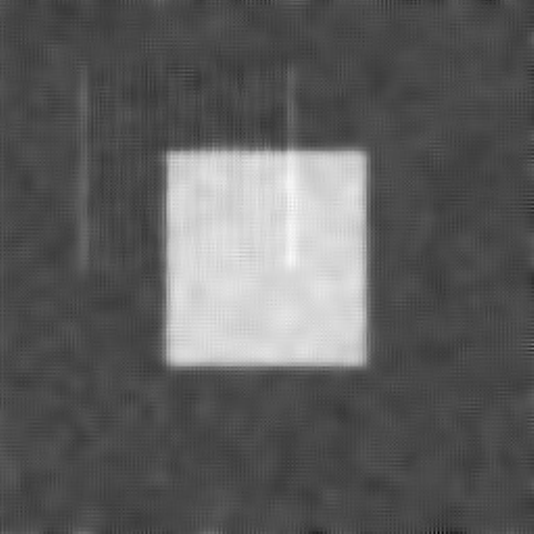} \\  
Noisy original image & Structures \\
\includegraphics[width=0.3\textwidth]{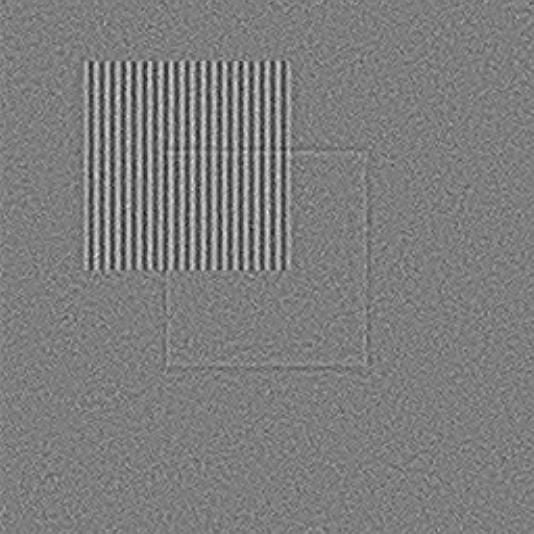} &
\includegraphics[width=0.3\textwidth]{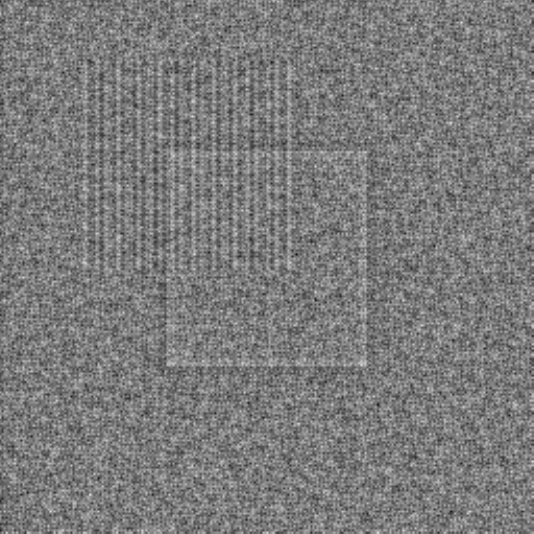} \\ 
Textures & Noise
\end{tabular}
\caption{Results given by $F_{\lambda,\mu,\delta}^{AC}$ applied on the synthetic image.}
\label{fig:accar}
\end{figure}

\begin{figure}[t!]
\hspace{-7mm}
\begin{tabular}{ccc}
\includegraphics[width=0.33\textwidth]{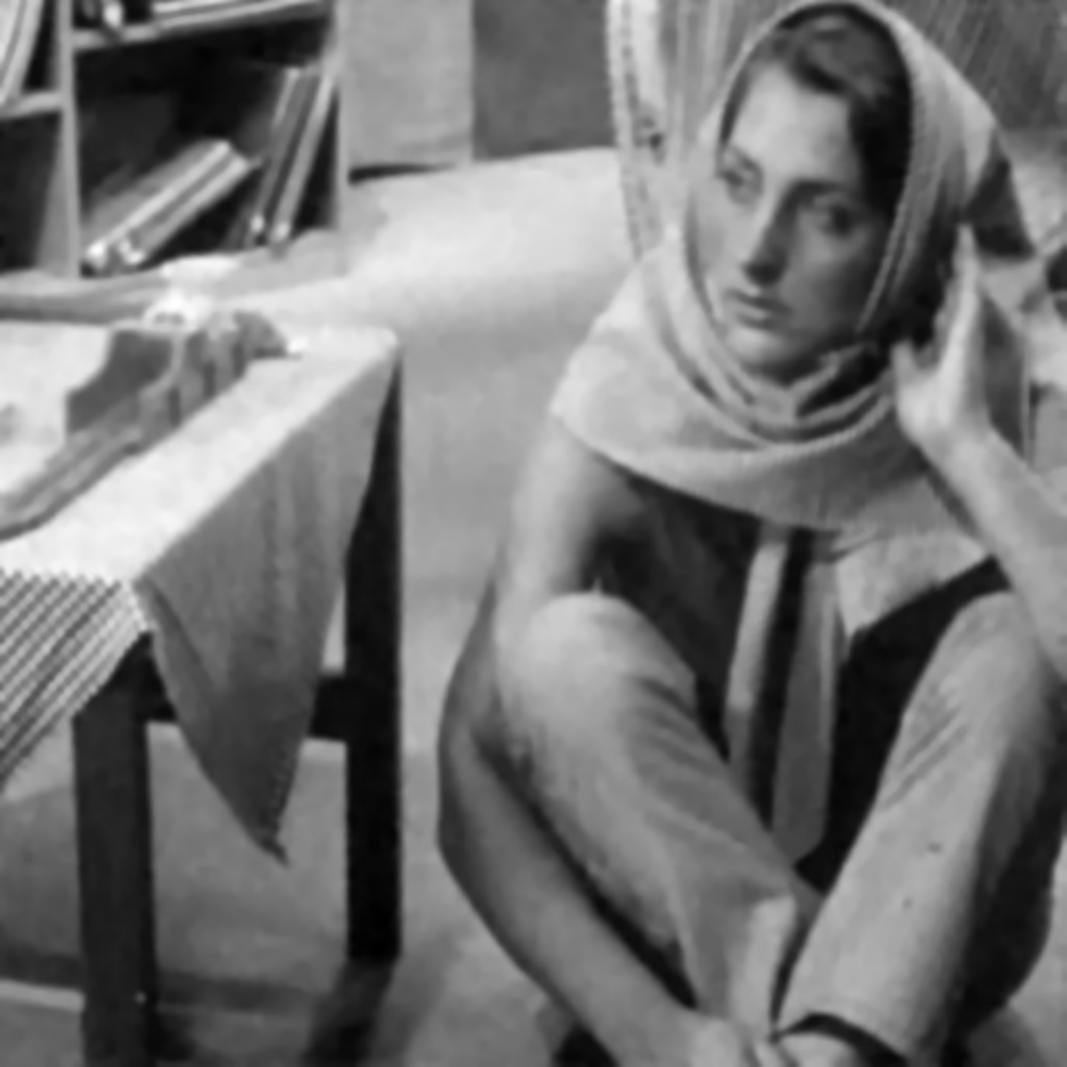} &  
\includegraphics[width=0.33\textwidth]{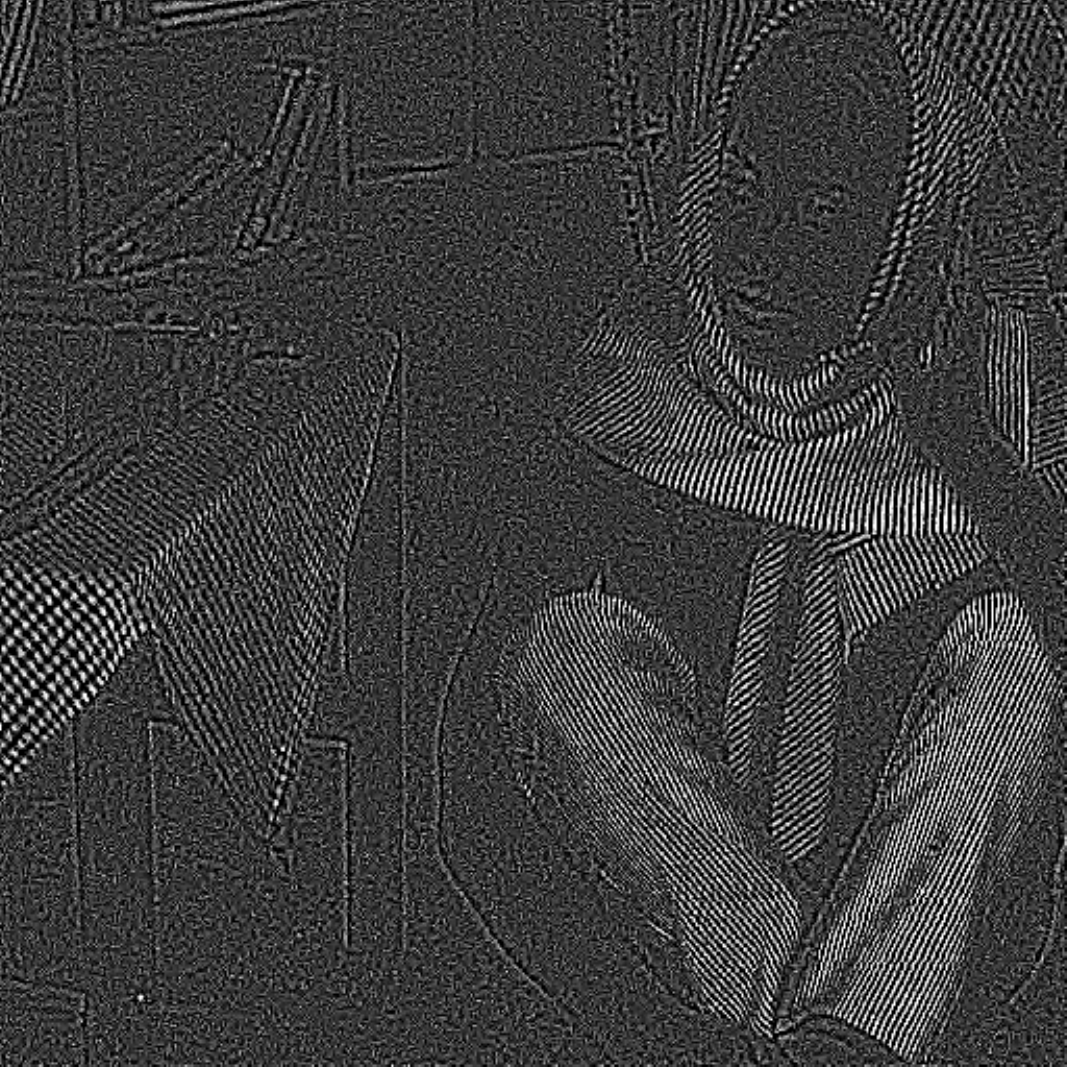} &
\includegraphics[width=0.33\textwidth]{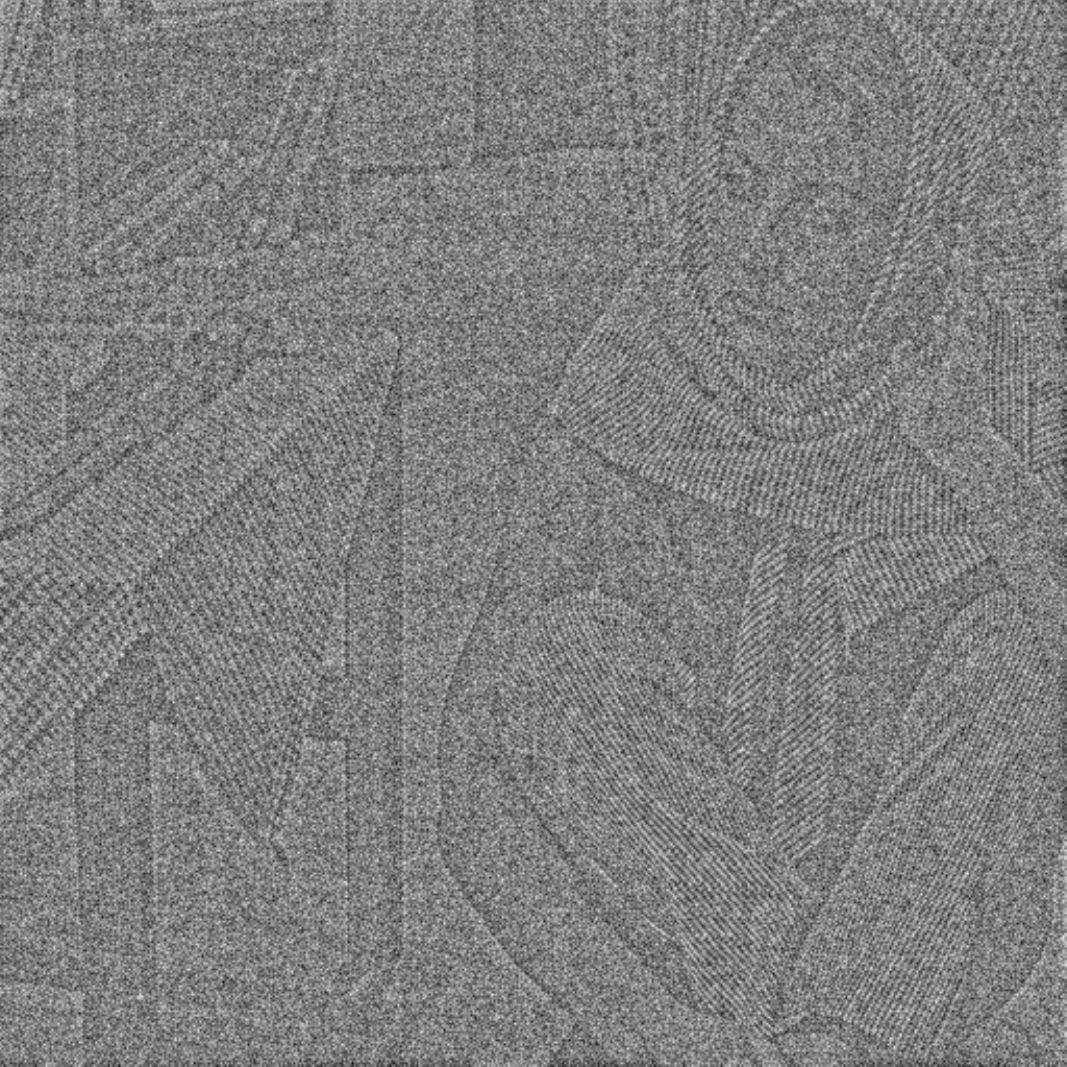} \\ 
Structures & Textures & Noise \\

\includegraphics[width=0.33\textwidth]{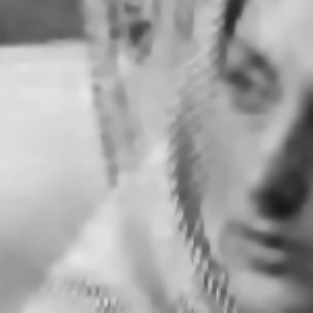} &  \includegraphics[width=0.33\textwidth]{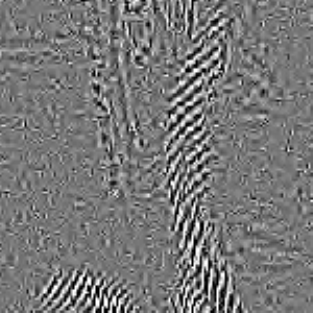} &
\includegraphics[width=0.33\textwidth]{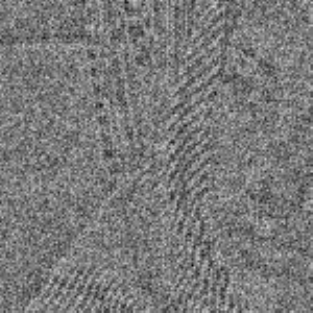} \\
Structures zoom & Textures zoom & Noise zoom

\end{tabular}
\caption{Results given by $F_{\lambda,\mu,\delta}^{AC}$ applied on Barbara.}
\label{fig:ac}
\end{figure}

\begin{figure}[t!]
\hspace{-7mm}
\begin{tabular}{ccc}
\includegraphics[width=0.49\textwidth]{batiment_noise} &  
\includegraphics[width=0.49\textwidth]{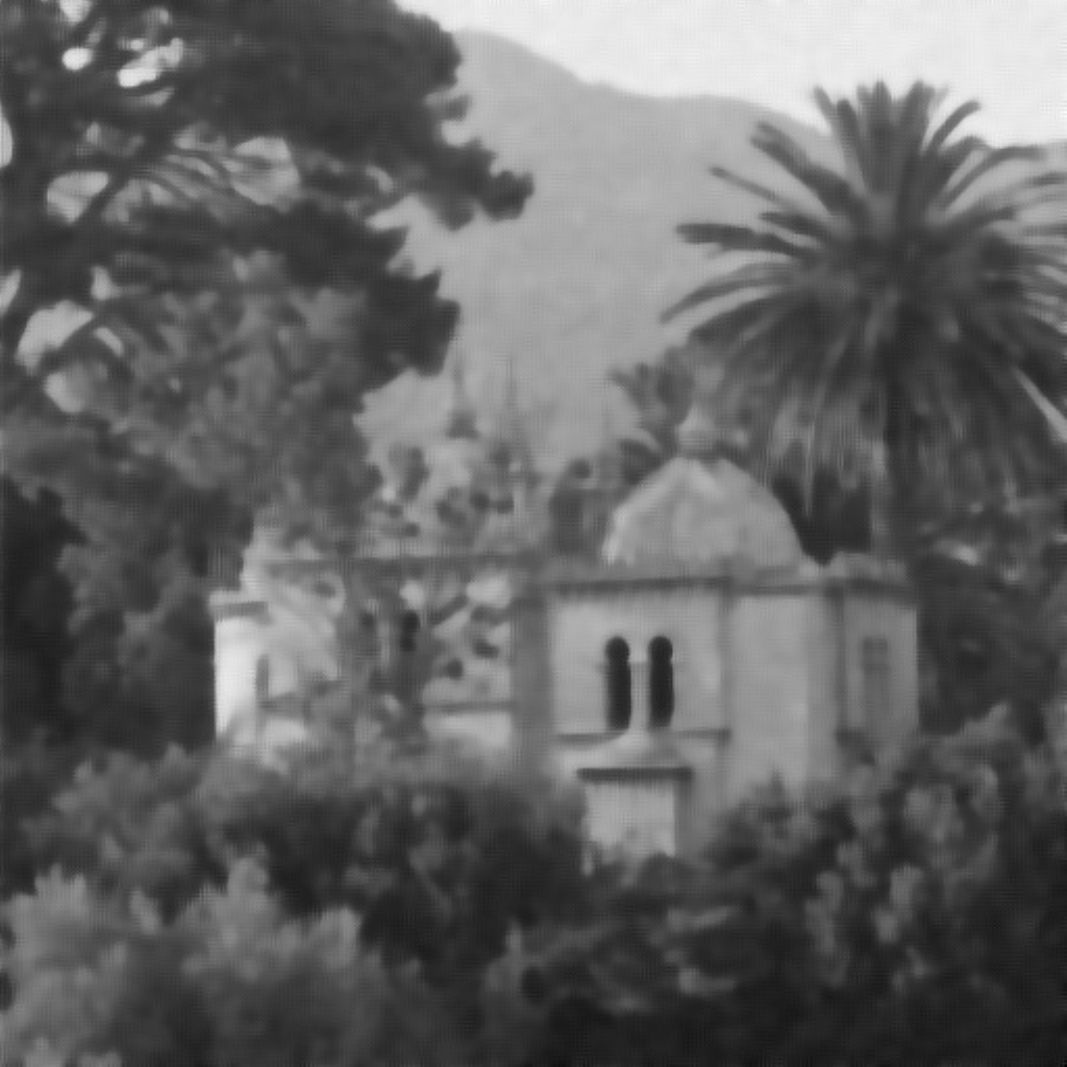} \\  
Noisy original image & Structures \\
\includegraphics[width=0.49\textwidth]{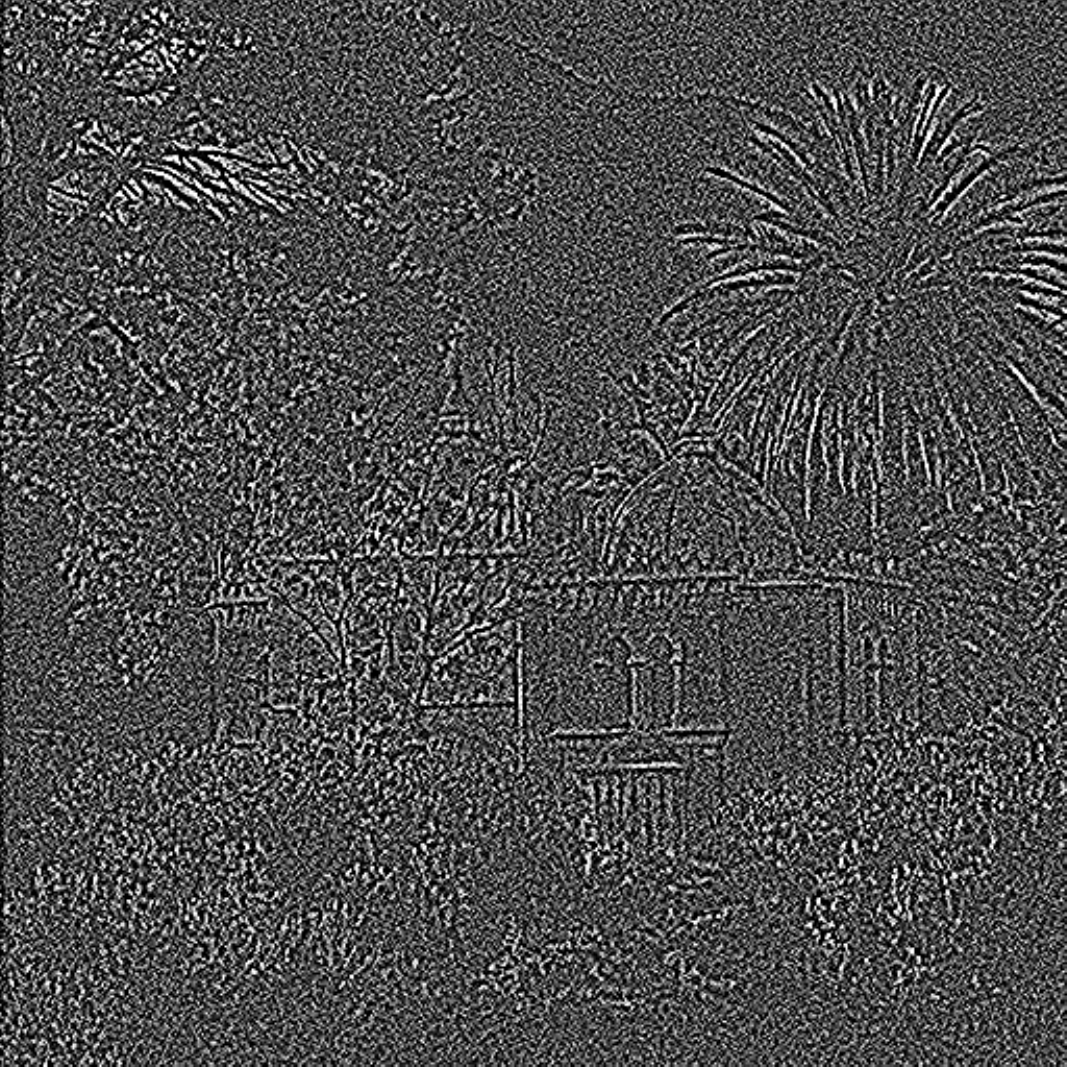} &
\includegraphics[width=0.49\textwidth]{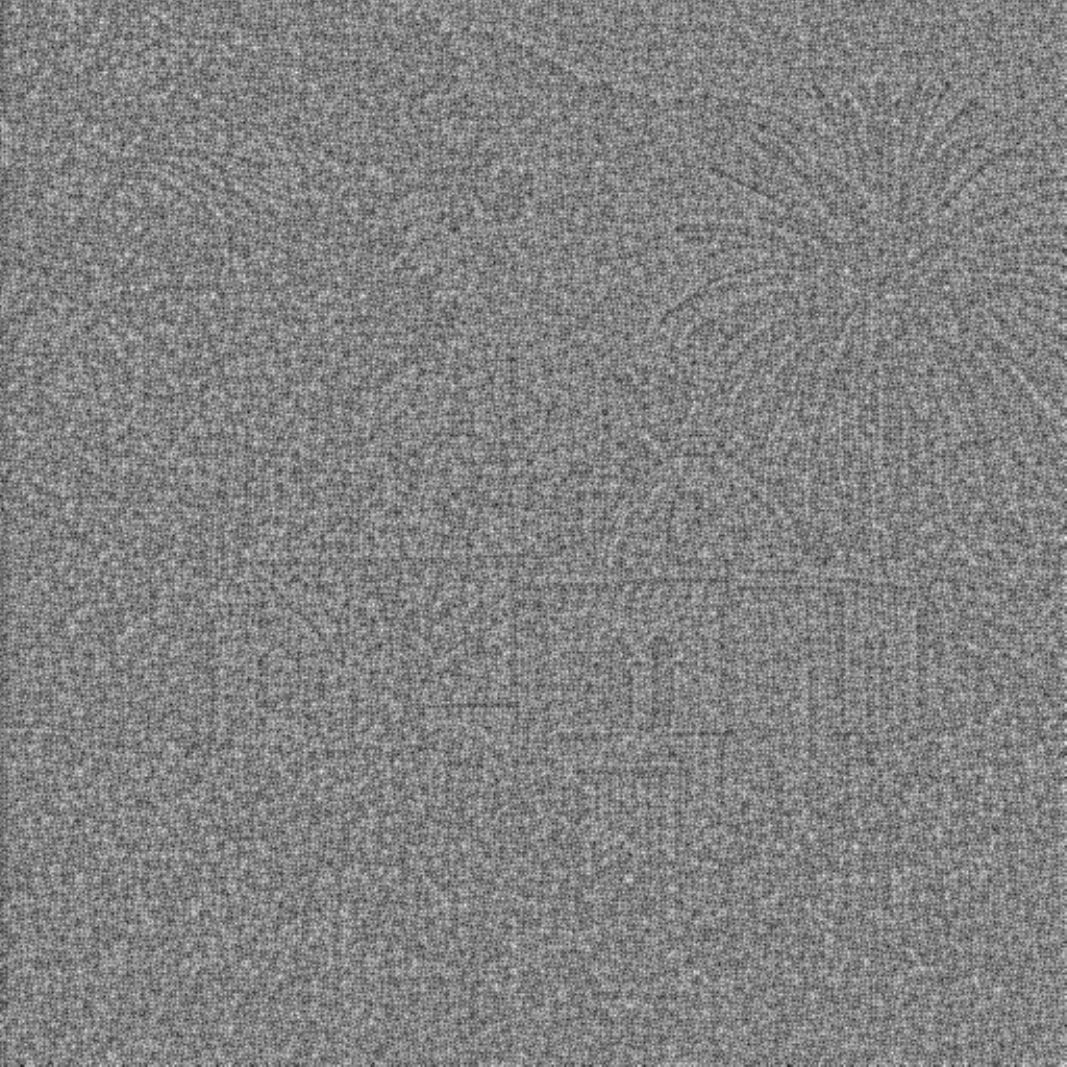} \\ 
Textures & Noise
\end{tabular}
\caption{Results given by $F_{\lambda,\mu,\delta}^{AC}$ applied on the outdoor.}
\label{fig:acbat}
\end{figure}

Figures \ref{fig:accar}, \ref{fig:ac}, \ref{fig:acbat} give the results obtained with this model on the same three image test used in the previous section. The parameters are set to 
\begin{itemize}
\item Synthetic: $\mu=1000$, $\lambda=20$, $\sigma=20$, $\eta=0.2$,
\item Barbara: $\mu=100$, $\lambda=1$, $\sigma=20$, $\eta=0.1$,
\item Outdoor: $\mu=5000$, $\lambda=20$, $\sigma=20$, $\eta=0.1$.
\end{itemize}

We can see that the textures are better denoised by this model. This is a consequence of a better noise modeling by distributions in the Besov space. But the residual texture is more important than the one given by our algorithm in the noise part. This behavior is certainly implied by the local adaptivity capabilities of our algorithm. In the next section, we propose to merge the use of Besov space and the local adpativity in a unique algorithm.

\section{Adaptative wavelet algorithm}\label{sec:adaptwav}
We saw in the previous sections that the Besov space $\Dot{B}^{\infty}_{-1,\infty}$ is better adapted to modelize the noise than the set $G_{\mu}$ (where $\mu$ is very small). In the same way, we saw that local adaptivity implies a decrease of the residual quantity of texture in the noise part. So the question is: how to merge these advantages in a unique algorithm? We propose the following model as an answer:

\begin{equation}\label{eqn:uvwg1}
F_{\lambda,\mu,\delta}^{JG2}(u,v,w)=J(u)+J^*\left(\frac{v}{\mu}\right)+B^*\left(\frac{w}{\delta}\right)+(2\lambda)^{-1}\|f-u-\nu_1 v-\nu_2 w\|_{L^2}^2,
\end{equation}

where $u\in BV, v\in G, w\in E_{\delta}$ (as defined in the previous section) and the $\nu_i$ functions are the ones defined in section \ref{sec:uvwlocal}. A minimizer is given by the following proposition.

\begin{proposition}\label{pro:uvwjg2}
Let $u\in BV$, $v\in G_{\mu}$, $w\in E_{\delta}$ be the structures, textures and noise parts respectively and $(\nu_1(f)(x,y),\nu_2(f)(x,y))$ be defined as in proposition \ref{prop:uvw}. Then a minimizer of:
\begin{equation}
(\hat{u},\hat{v},\hat{w})\in\underset{(u,v,w)\in BV\times G_{\mu}\times E_{\delta}}{\arg} \min F_{\lambda,\mu,\delta}^{JG2}(u,v,w)
\end{equation}
is given by:
\begin{align*}
\hat{u}&=f-\nu_1\hat{v}-\nu_2\hat{w}-P_{G_{\lambda}}(f-\nu_1\hat{v}-\nu_2\hat{w}), \\
\hat{v}&=P_{G_{\mu}}\left(\frac{f-\hat{u}-\nu_2\hat{w}}{\nu_1+\kappa}\right), \\
\hat{w}&=\frac{f-\hat{u}-\nu_1\hat{v}}{\nu_2+\kappa}-\frac{\lambda}{\delta\nu_2^2+\kappa}\hat{WST}\left(\frac{\delta \nu_2}{\lambda}\left(f-\hat{u}-\nu_1\hat{v}\right);\frac{2\delta^2\tilde{\nu}_2^2}{\lambda}\right),
\end{align*}
where $P_{G_{\lambda}}$ is the Chambolle's nonlinear projector defined in \cite{chambolle} and $\hat{WST}(f,\delta)$ is a modified Wavelet Soft Thresholding of $f$. The threshold is dependent on the location in each wavelet coefficient, $\tilde{\nu_2}$ is a pyramidal version of $\nu_2$ (see fig \ref{fig:nures}).\\
\end{proposition}

A proof of this proposition is given in appendix \ref{ann:b}.\\

\begin{figure}[t!]
\centering
\begin{tabular}{cc}
\includegraphics[width=0.33\textwidth]{nu} &  \includegraphics[width=0.33\textwidth]{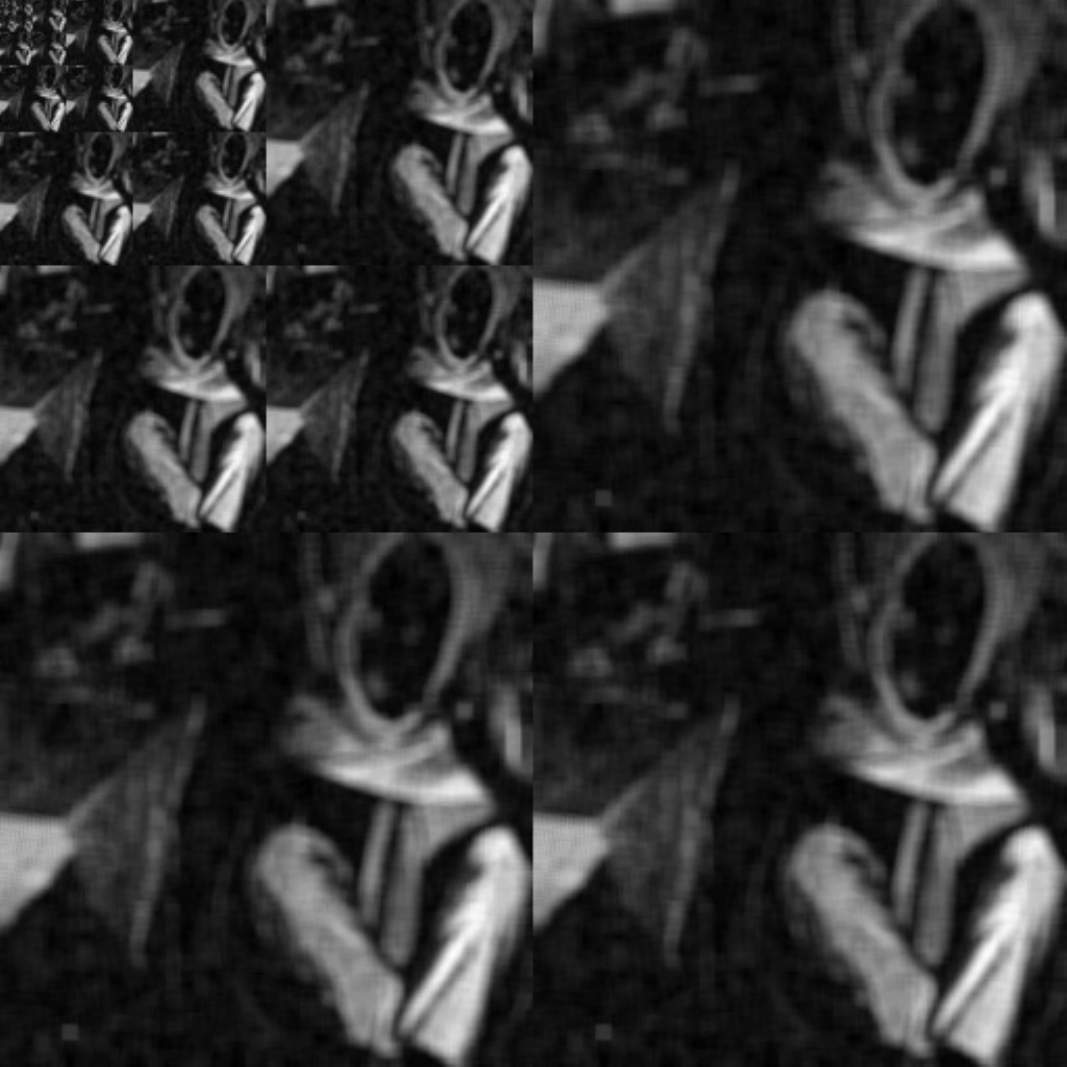} \\
\end{tabular}
\caption{Local smoothed partition $\nu$ and it's pyramidal version $\tilde{\nu}$.}
\label{fig:nures}
\end{figure}

\begin{figure}[t!]
\centering
\begin{tabular}{ccc}
\includegraphics[width=0.3\textwidth]{carre_texture_noisy} &  
\includegraphics[width=0.3\textwidth]{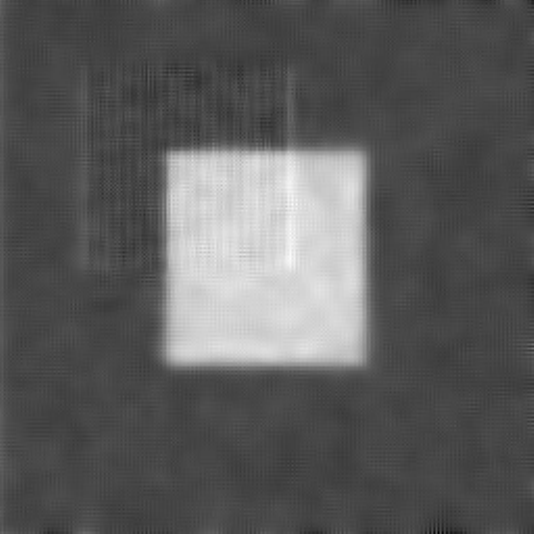} \\ 
Noisy original image & Structures \\ 
\includegraphics[width=0.3\textwidth]{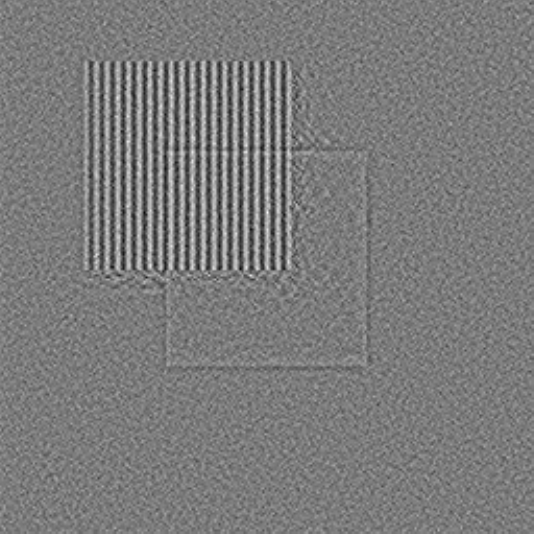} &
\includegraphics[width=0.3\textwidth]{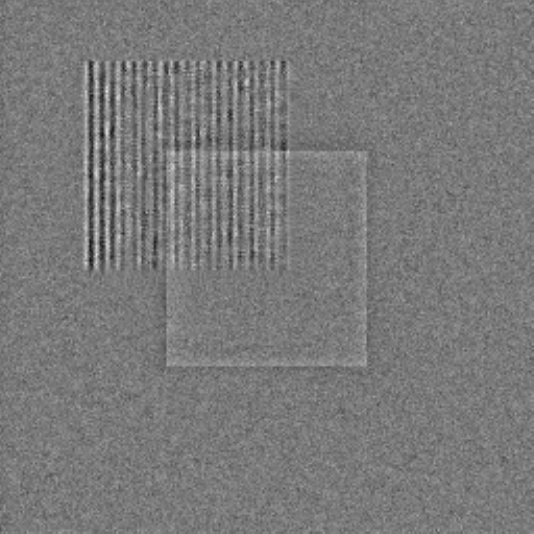} \\ 
Textures & Noise
\end{tabular}
\caption{Results given by $F_{\lambda,\mu,\delta}^{JG2}(u,v,w)$ applied on the synthetic image.}
\label{fig:jg2car}
\end{figure}

\begin{figure}[t!]
\hspace{-7mm}
\begin{tabular}{ccc}
\includegraphics[width=0.33\textwidth]{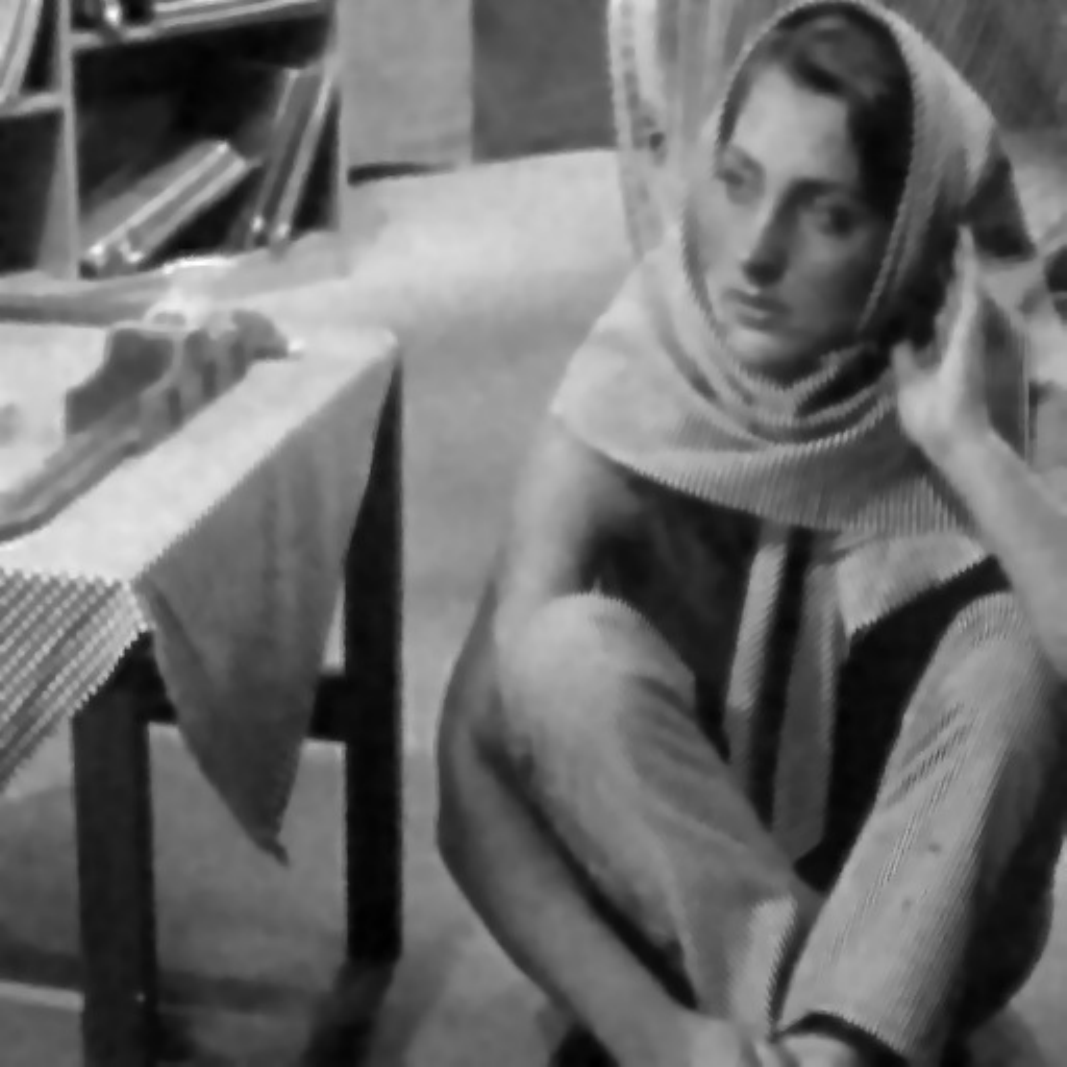} &
\includegraphics[width=0.33\textwidth]{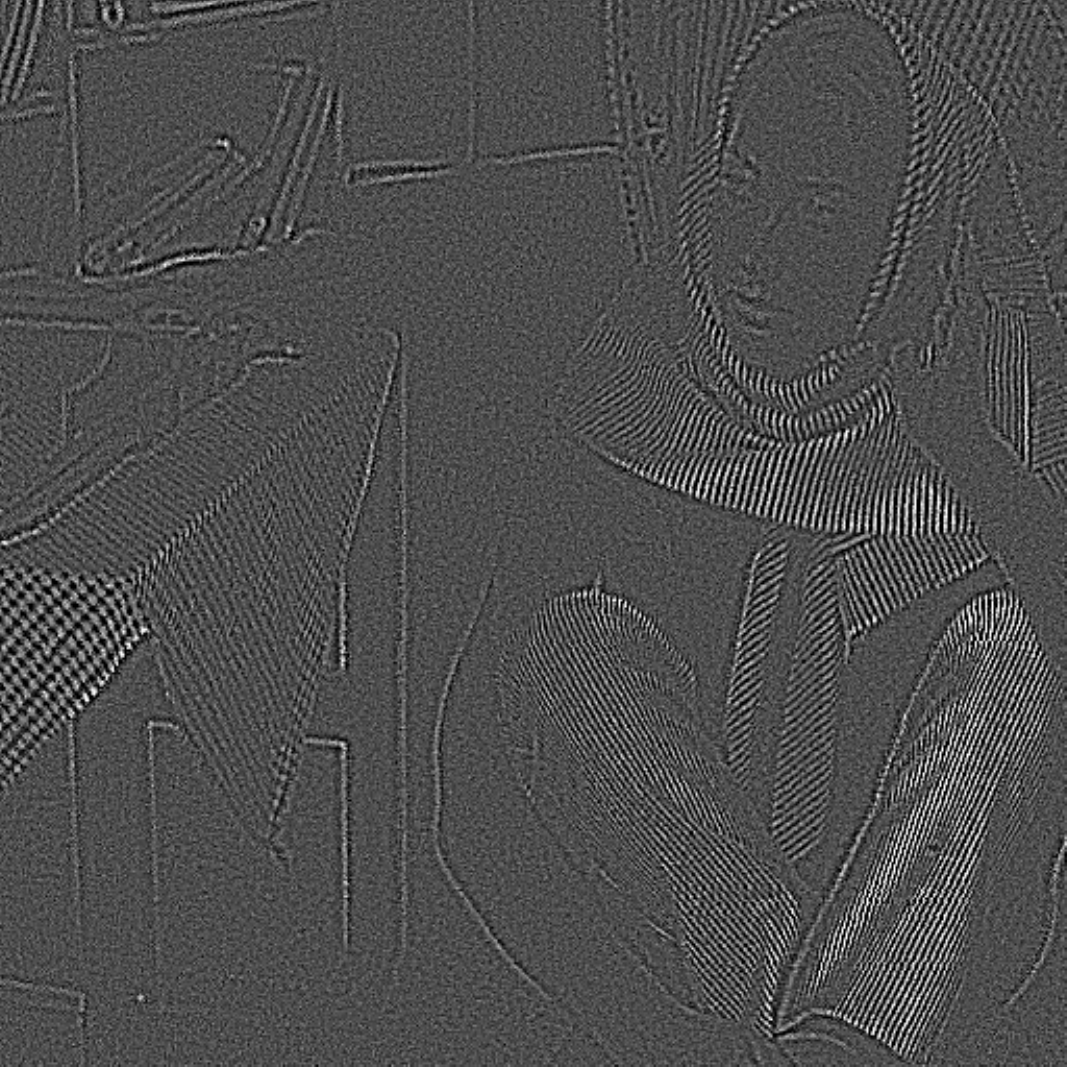} &  
\includegraphics[width=0.33\textwidth]{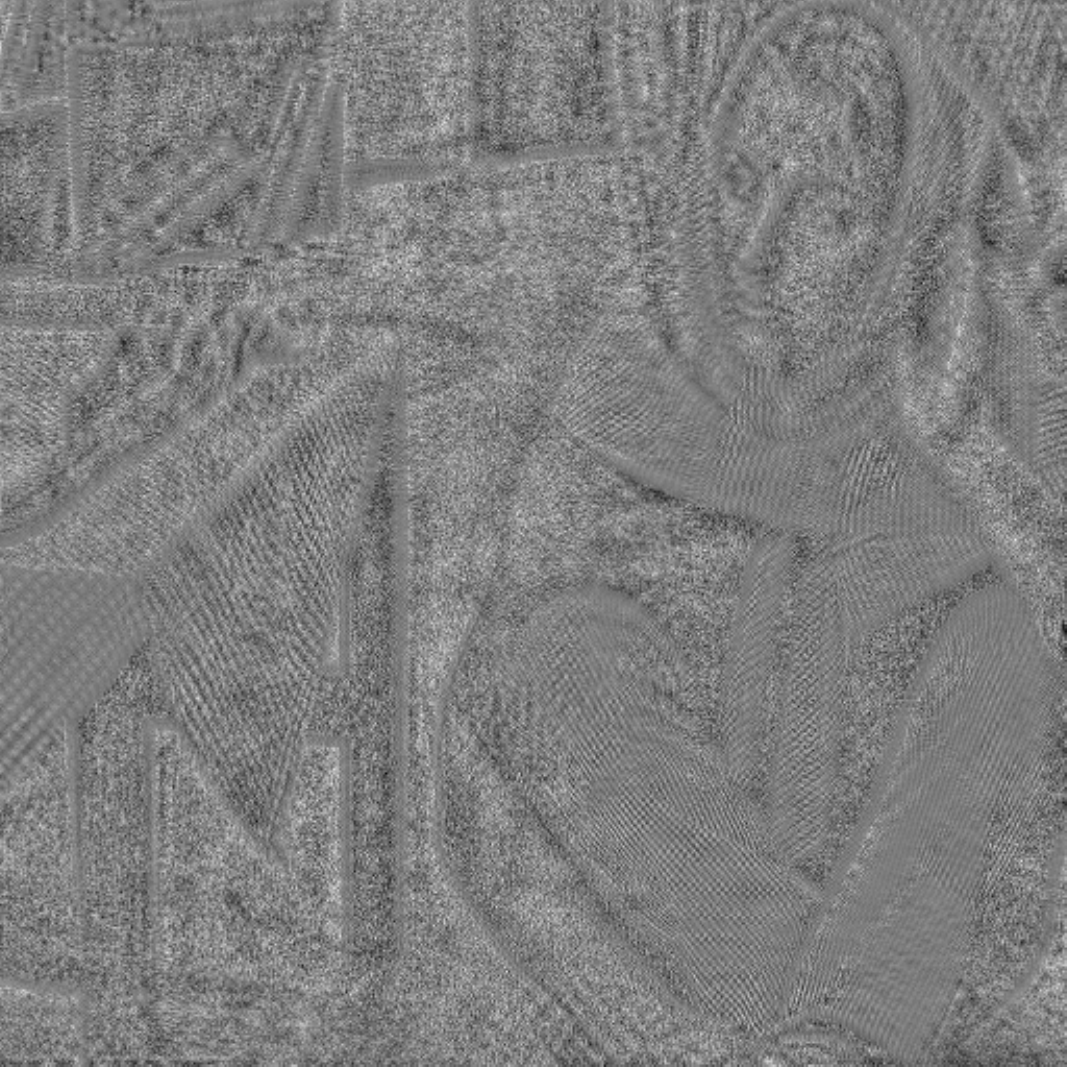} \\
Structures & Textures & Noise \\
\includegraphics[width=0.33\textwidth]{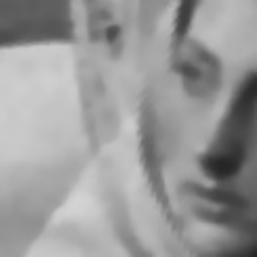} &
\includegraphics[width=0.33\textwidth]{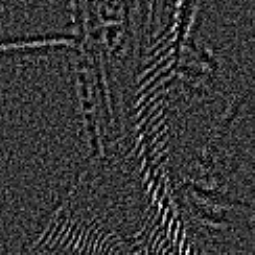} &  \includegraphics[width=0.33\textwidth]{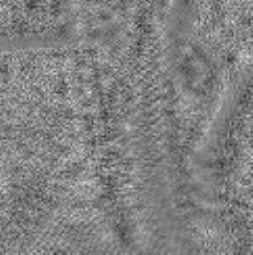} \\
Structures zoom & Textures zoom & Noise zoom

\end{tabular}
\caption{Results given by $F_{\lambda,\mu,\delta}^{JG2}(u,v,w)$ applied on Barbara.}
\label{fig:uvwjg2}
\end{figure}

\begin{figure}[t!]
\hspace{-7mm}
\begin{tabular}{ccc}
\includegraphics[width=0.49\textwidth]{batiment_noise} &  
\includegraphics[width=0.49\textwidth]{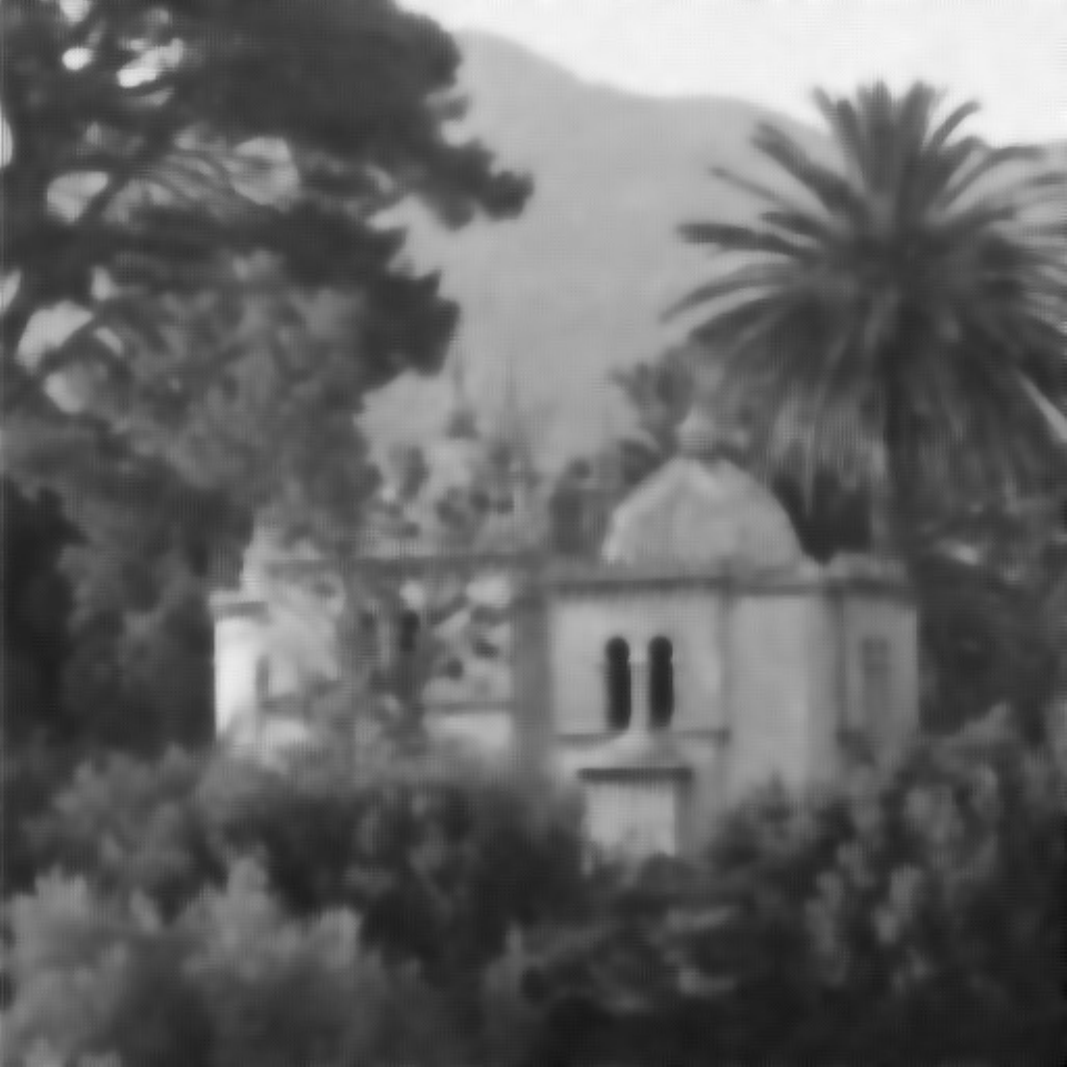} \\  
Noisy original image & Structures \\
\includegraphics[width=0.49\textwidth]{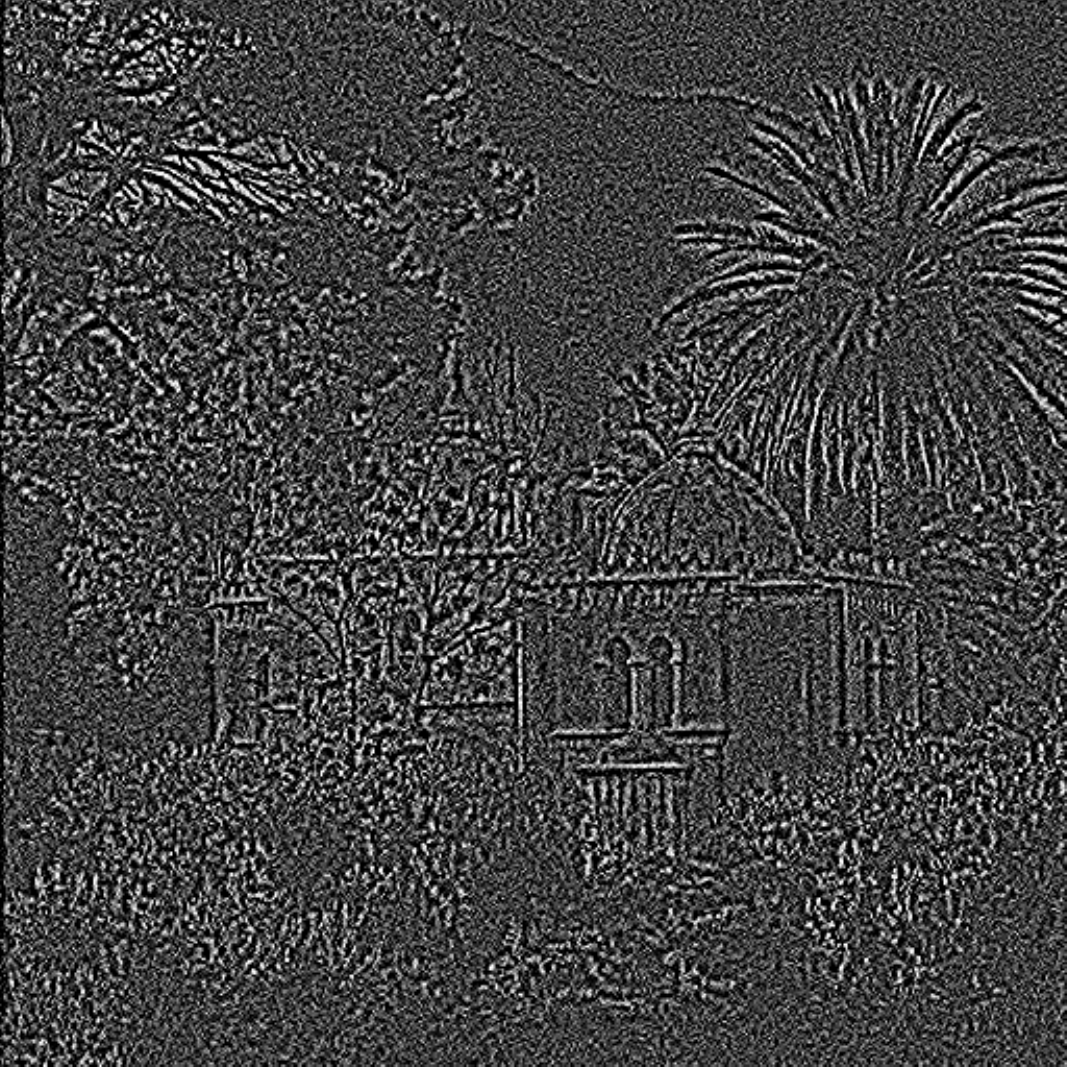} &
\includegraphics[width=0.49\textwidth]{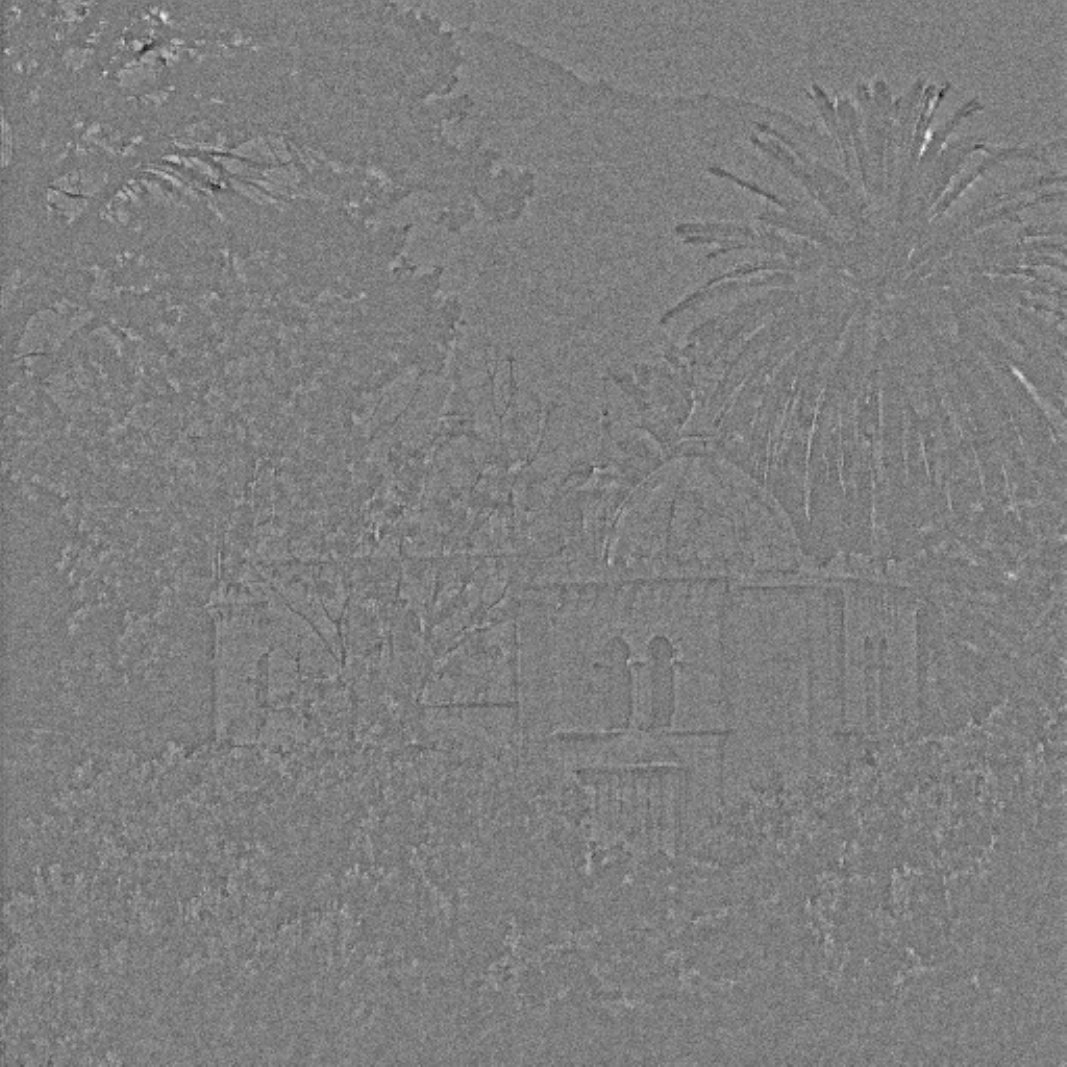} \\ 
Textures & Noise
\end{tabular}
\caption{Results given by $F_{\lambda,\mu,\delta}^{JG2}(u,v,w)$ applied on the outdoor image.}
\label{fig:jg2bat}
\end{figure}

This ``merged-algorithm'' gives the results showed in figure \ref{fig:jg2car}, \ref{fig:uvwjg2}.and \ref{fig:jg2bat}.  The parameters are set to 
\begin{itemize}
\item Synthetic: $\mu=1000$, $\lambda=50$, $\sigma=20$, $\eta=0.7$, window's size for $\nu_i=7$,
\item Barbara: $\mu=1000$, $\lambda=30$, $\sigma=20$, $\eta=0.5$, window's size for $\nu_i=7$,
\item Outdoor: $\mu=1000$, $\lambda=30$, $\sigma=20$, $\eta=0.3$, window's size for $\nu_i=5$.
\end{itemize}

As we expected, the combination of the two properties (Besov space and local adaptivity) increases the quality of the decomposition. Textures are better denoised and the noise part contains less residual texture. However, the noise is ``damaged'' by the local behaviour of the algorithm. 

\section{Conclusion}
\label{sec:conclusion}

In this paper, we presented an original extension of the two components image decomposition algorithm of Aujol et al \cite{aujol} in order to deal with the case of noisy images. This new algorithm uses the assumption that noise is a highly oscillatory function and uses a local adaptivity principle. We compared our results with ones obtained by the Aujol-Chambolle's algorithm and we then proposed a ``merged-algorithm'' which incorporates the advantages of the two preceding models. The results are quite promising.
Futur work will focus on the choice of the different parameters. This choice is still not automated, and a manual choice of the good parameters is not an easy task. Another line of research is on the choice of the functions $\nu_i$. Another modeling of the local behavior of the image may be obtained from a local approximation of oscillation quantity in term of the $G-$norm. Some other interesting works are done in \cite{haddad,haddad2,triet,triet2,lieu} where different functional spaces are tested and may be an alternative way to the space $G$.
It is another important issue to built objective comparison criteria, for comparing the different decomposition algorithms. We developped metrics adapted to each image component. These metrics will be describe in a futur paper.

\section*{Acknowledgements}
The author would like to thank Prof. Yves Meyer for his salient remarks in the redaction of this paper. We also wants to thank the reviewers for their remarks which resulted in a great improvement of the paper.
\appendix
\section{Appendix A}
\label{ann:a}
The proof of proposition \ref{prop:uvw} is largely inspired from \cite{chambolle}, but is detailed now for the reader convenience.\\ The Euler-Lagrange equations with respect to $u$ are

\begin{eqnarray}
-\frac{1}{\lambda}(f-u-\nu_1 v-\nu_2 w)+\partial J(u) \ni 0\\
\Leftrightarrow u\in \partial J^*\left(\frac{1}{\lambda}(f-u-\nu_1 v-\nu_2 w)\right).
\end{eqnarray}
By adding $f-u-\nu_1 v-\nu_2 w$ on each side and multiplying by $\frac{1}{\lambda}$, it is deduced:
\begin{eqnarray}
\frac{1}{\lambda}(f-\nu_1 v-\nu_2 w)\in & \frac{1}{\lambda}(f-u-\nu_1 v-\nu_2 w)+ \\ \nonumber
& \frac{1}{\lambda}\partial J^*\left(\frac{1}{\lambda}(f-u-\nu_1 v-\nu_2 w)\right).
\end{eqnarray}
By setting $\eta=(f-u-\nu_1 v-\nu_2 w)/\lambda$, it comes:
\begin{equation}
\frac{1}{\lambda}(f-\nu_1 v-\nu_2 w)\in \eta+\frac{1}{\lambda}\partial J^*(\eta).
\end{equation}
Thus, using the results in \cite{chambolle}, we get
\begin{equation}
\eta=P_G\left(\frac{1}{\lambda}(f-\nu_1 v-\nu_2 w)\right)=\frac{1}{\lambda}P_{G_{\lambda}}(f-\nu_1 v-\nu_2 w).
\end{equation}
Using the expression of $\eta$ we conclude that
\begin{equation}
\hat{u}=f-\nu_1v-\nu_2w-P_{G_{\lambda}}(f-\nu_1v-\nu_2w). 
\end{equation}

The $u$ part is now completed. To compute the $v$ and $w$ parts, we use the following lemma.
\begin{lemme}\label{lem:uvw}
Let be $f\in L^2(\R^2)$, $v\in G_{\mu}$ and $\nu:\R^2\rightarrow ]0;1[$. As $\nu$ is more regular than $v$, we assume $\nu$ as locally constant compared to $v$ (this assumption is like a first-order assumption). Then the solution:
\begin{equation}\label{eq:lemv}
\hat{v}=\underset{v\in G_{\mu}}{\arg} \inf\left\{(2\lambda)^{-1}\|f-\nu v\|_{L^2}^2+J^*\left(\frac{v}{\mu}\right)\right\}
\end{equation}
can be approximated by:
\begin{equation}
\hat{v}=P_{G_{\mu}}\left(\frac{f}{\nu}\right)
\end{equation}
\end{lemme}

By setting $\eta=\frac{v}{\mu}$, we get 
\begin{equation}
F(\eta)=(2\lambda)^{-1}\|f-\mu\nu \eta\|_{L^2}^2+J^*\left(\eta\right)
\end{equation}
and now we seek for
\begin{equation}\label{eq:lem2}
\hat{\eta}=\underset{\eta}{\arg} \inf F(\eta).
\end{equation}

First, let's show that it exists some $F_m(\eta)$ and $F_M(\eta)$ such that
\begin{equation}
F_m(\eta)\leqslant F(\eta) \leqslant F_M(\eta).
\end{equation}

Let's begin by $F_M(\eta)$,
\begin{align}
F(\eta)&=(2\lambda)^{-1}\left\|\mu\nu\left(\frac{f}{\mu\nu}-\eta\right)\right\|_{L^2}^2+J^*(\eta)\\
&\leqslant (2\lambda)^{-1}\|\mu\nu\|_{L^2}^2\left\|\frac{f}{\mu\nu}-\eta\right\|_{L^2}^2+J^*(\eta)
\end{align}

then we choose
\begin{equation}
F_M(\eta)=(2\lambda)^{-1}\|\mu\nu\|_{L^2}^2\left\|\frac{f}{\mu\nu}-\eta\right\|_{L^2}^2+J^*(\eta).
\end{equation}

Now let's deal with $F_m(\eta)$, if we choose
\begin{equation}
F_m(\eta)=(2\lambda)^{-1}\frac{\mu}{\|\nu\|_{L^2}^2}\left\|\frac{f}{\mu\nu}-\eta\right\|_{L^2}^2+J^*(\eta).
\end{equation}

We have
\begin{equation}
\left\|\frac{f}{\mu\nu}-\eta\right\|_{L^2}^2\leqslant \|\nu\|_{L^2}^2\left\|\frac{1}{\nu}\left(\frac{f}{\mu\nu}-\eta\right)\right\|_{L^2}^2
\end{equation}
then
\begin{align}
\Longrightarrow \frac{1}{\|\nu\|_{L^2}^2}\left\|\frac{f}{\mu\nu}-\eta\right\|_{L^2}^2&\leqslant\left\|\frac{1}{\nu}\left(\frac{f}{\mu\nu}-\eta\right)\right\|_{L^2}^2\\
&=\left\| \frac{1}{\mu\nu^2}\left( f-\mu\nu\eta \right) \right\|_{L^2}^2\\
&\leqslant \left\| \frac{1}{\mu\nu^2} \right\|_{L^2}^2\|f-\mu\nu\eta\|_{L^2}^2.
\end{align}
We suppose that $\left\|\frac{1}{\nu^2} \right\|_{L^2}^2\lesssim 1$ (we can verify it in all our experiments). Then
\begin{equation}
\frac{1}{\|\nu\|_{L^2}^2}\left\|\frac{f}{\mu\nu}-\eta\right\|_{L^2}^2 \leqslant \frac{1}{\mu}\|f-\mu\nu\eta\|_{L^2}^2.
\end{equation}
This implies
\begin{equation}
F_m(\eta)\leqslant F(\eta)
\end{equation}

Now we want to characterize $F(\eta)$ by studying the behaviour of $F_m(\eta)$ and $F_M(\eta)$. First, as the three energy are convex, we have
\begin{equation}
\inf F_m(\eta)\leqslant\inf F(\eta) \leqslant \inf F_M(\eta).
\end{equation}
Let calculate $F_m^{min}=\inf F_m(\eta)$ and $F_M^{min}=\inf F_M(\eta)$. First, we have
\begin{equation}
\hat{\eta}_M = \underset{\eta}{\arg} \inf F_M(\eta).
\end{equation}
\begin{equation}
\Longleftrightarrow -(2\lambda)^{-1}\|\mu\nu\|_{L^2}^2\left( \frac{f}{\mu\nu} -\eta\right)+\partial J^*(\eta)\ni 0
\end{equation}
\begin{equation}
\Longleftrightarrow \eta+\frac{2\lambda}{\|\mu\nu\|_{L^2}^2}\partial J^*(\eta) \ni \frac{f}{\mu\nu}
\end{equation}
Chambolle \cite{chambolle} shows that
\begin{equation}
\hat{\eta}_M=P_{G_1}\left( \frac{f}{\mu\nu} \right)=\frac{1}{\mu}P_{G_{\mu}}\left(\frac{f}{\nu} \right)
\end{equation}
and then
\begin{equation}
F_M^{min}=(2\lambda)^{-1}\|\nu\|_{L^2}^2\left\| \frac{f}{\nu}-P_{G_{\mu}}\left(\frac{f}{\nu} \right) \right\|_{L^2}^2.
\end{equation}
By the same way, we find that
\begin{equation}
\hat{\eta}_m=P_{G_1}\left( \frac{f}{\mu\nu} \right)=\frac{1}{\mu}P_{G_{\mu}}\left(\frac{f}{\nu} \right)
\end{equation}
and
\begin{equation}
F_m^{min}=\frac{(2\lambda)^{-1}}{\|\nu\|_{L^2}^2}\left\| \frac{f}{\nu}-P_{G_{\mu}}\left(\frac{f}{\nu} \right) \right\|_{L^2}^2.
\end{equation}
Then if we set $B=(2\lambda)^{-1}\left\| \frac{f}{\nu}-P_{G_{\mu}}\left(\frac{f}{\nu} \right) \right\|_{L^2}^2$, we get
\begin{equation}
\frac{B}{\|\nu\|_{L^2}^2}\leqslant F^{min}=F(\hat{\eta})\leqslant B\|\nu\|_{L^2}^2
\end{equation}
Remind that we suppose that $\left\|\frac{1}{\nu^2} \right\|_{L^2}^2\lesssim 1$, in pratice we can easily check that $\|\nu\|_{L^2}\approx 1$ and as $\hat{\eta}_m=\hat{\eta}_M$, a good approximation of $\hat{\eta}$ is given by
\begin{equation}
\hat{\eta}=\frac{1}{\mu}P_{G_{\mu}}\left(\frac{f}{\nu}\right)
\end{equation}

%
Since $\hat{\eta}=\frac{\hat{v}}{\mu}$, we conclude the proof of the lemma that $\hat{v}$ is well approximated by:
\begin{equation}
\hat{v}=P_{G_{\mu}}\left(\frac{f}{\nu}\right).
\end{equation}

We apply this lemma two times to $F_{\lambda ,\mu_1 ,\mu_2}^{JG}(u,v,w)$. Then $\hat{v}$ is obtained from $u$ and $w$, while $\hat{w}$ is obtained from $u$ and $v$. This concludes the proof of proposition \ref{prop:uvw}.

\section{Appendix B}
\label{ann:b}

Now, we give the proof of proposition \ref{pro:uvwjg2} of section \ref{sec:adaptwav}. First, it is easy to see that for the components $u$ and $v$, the proof is the same as the one of proposition \ref{prop:uvw} given in appendix \ref{ann:a}. So we will focus on $w$ and then suppose the components $u$ and $v$ are fixed. We need to find
\begin{equation}\label{eqn:propbesov}
\hat{w}=\underset{w\in E_{\delta}}{\arg}\min\left\{ (2\lambda)^{-1}\left\|f-u-\nu_1v-\nu_2w\right\|_{L^2}^2+B^{*}\left(\frac{w}{\delta}\right)\right\}.
\end{equation}
To get this solution, the following lemma will be used
\begin{lemme}
Let $f \in L^2$, $w\in E_{\delta}$ and $\nu:\mathbb{R}^2 \rightarrow ]0;1[$. Assume that, the function $\nu$ can be considered as locally constant compared to the variation of $v$. Then the minimizer:
\begin{equation}\label{eqn:lembesov}
\hat{w}=\underset{w\in E_{\delta}}{\arg}\min\left\{(2\lambda)^{-1}\|f-\nu w\|_{L^2}^2+B^{*}\left(\frac{w}{\delta}\right)\right\}
\end{equation}
is given by:
\begin{equation}
\hat{w}=\frac{f}{\nu}-\frac{\lambda}{\delta\nu^2}WST\left(\frac{\delta f\nu}{\lambda};\frac{2\delta^2 \tilde{\nu}^2}{\lambda}\right),
\end{equation}.
\end{lemme}
where $WST(f,\delta)$ is the Wavelet Soft Thresholding operator (see \cite{chambolle2,donohoseuil}).

The proof is based on the differentiation of (\ref{eqn:lembesov}) with respect to $w$. Using the Euler-Lagrange equation, we get
\begin{eqnarray}
0\in -\frac{\nu}{\lambda}(f-\nu w)+\frac{1}{\delta}\partial B^{*}\left(\frac{w}{\delta}\right)\\
\Leftrightarrow \frac{w}{\delta}\in \partial B\left(\frac{\delta\nu (f-\nu w)}{\lambda}\right).
\end{eqnarray}
Let $\eta=\frac{\delta \nu (f-\nu w)}{\lambda}$. Then 
\begin{eqnarray}
0\in \frac{\lambda}{\delta^2 \nu^2}\eta -\frac{f}{\delta \nu}+\partial B(\eta)\\
\Leftrightarrow 0\in \eta -\frac{\delta f\nu}{\lambda}+\frac{\delta^2 \nu^2}{\lambda}\partial B(\eta).
\end{eqnarray}
This comes from the Euler-Lagrange equation of
\begin{equation}
\underset{\eta \in L^2}{\arg}\inf\left\{\frac{1}{2}\left\|\eta-\frac{\delta f\nu}{\lambda}\right\|_{L^2}^2+\frac{\delta^2 \nu^2}{\lambda}\|\eta\|_{B_{1,1}^1}\right\}.
\end{equation}
Based on the result of Chambolle et al. \cite{chambolle2} and because we suppose the $\nu$ behaves locally as a constant compared to $\eta$, we get
\begin{equation}
\hat{\eta}=WST\left(\frac{\delta f\nu}{\lambda};\frac{2\delta^2 \tilde{\nu}^2}{\lambda}\right)
\end{equation}
where $\tilde{\nu}$ is the pyramidal version of $\nu$ defined in section \ref{sec:adaptwav}. 
Now, we get from the expression of $\hat{\eta}$:
\begin{equation}
\frac{\delta f\nu}{\lambda}-\frac{\delta \nu^2 \hat{w}}{\lambda}=WST\left(\frac{\delta f\nu}{\lambda};\frac{2\delta^2\tilde{\nu}^2}{\lambda}\right)
\end{equation}

\begin{equation}
\Rightarrow \hat{w}=\frac{f}{\nu}-\frac{\lambda}{\delta \nu^2}WST\left(\frac{\delta f\nu}{\lambda};\frac{2\delta^2 \tilde{\nu}^2}{\lambda}\right).
\end{equation}

In order to end the proof of proposition \ref{pro:uvwjg2}, we just have to substitute $f$ by $f-u-\nu_1v$ and $\nu$ by $\nu_2$ in the previous lemma. Then we get:
\begin{equation}
\hat{w}=\frac{f-u-\nu_1v}{\nu_2}-\frac{\lambda}{\delta\nu_2^2}WST\left(\frac{\delta \nu_2}{\lambda}\left(f-u-\nu_1v\right);\frac{2\delta^2 \tilde{\nu}_2^2}{\lambda}\right).
\end{equation}

\nocite{*}

\bibliographystyle{IEEEbib}

\end{document}